\pdfoutput=1
\documentclass[10pt,twocolumn,letterpaper]{article}

\usepackage[pagenumbers]{wacv} 

\usepackage{graphicx}
\usepackage{amsmath}
\usepackage{amssymb}
\usepackage{booktabs}

\usepackage{pifont}
\usepackage{array}
\usepackage{caption}
\usepackage{subcaption}
\usepackage{enumitem}
\usepackage{multirow}

\newcommand{\hy}[1]{\textcolor{black}{#1}} 
\def\B#1{\textbf{#1}}
\newcommand{\cmark}{\ding{51}}%
\newcommand{\xmark}{\ding{55}}

\newcommand{\MulR}[2]{\multirow{#1}{*}{#2}} 
\newcommand{\MURot}[2]{\parbox[t]{2mm}{\MulR{#1}{\rotatebox[origin=c]{90}{#2}}}}

\def\B#1{\textbf{#1}}
\def\IT#1{\textit{#1}}

\def \su#1{\small{$^#1$}\large}

\usepackage{hyperref}
\hypersetup{pagebackref,breaklinks,colorlinks}

\usepackage[capitalize]{cleveref}
\crefname{section}{Sec.}{Secs.}
\Crefname{section}{Section}{Sections}
\Crefname{table}{Table}{Tables}
\crefname{table}{Tab.}{Tabs.}

\def\wacvPaperID{111} 
\def\confName{WACV}
\def\confYear{2024}

\begin{document}

\title{Multimodal Channel-Mixing: Channel and Spatial Masked AutoEncoder on Facial Action Unit Detection}

\author{\parbox{16cm}{\centering
    {\large Xiang Zhang\su{1}\hspace{5mm} Huiyuan Yang\su{2}\hspace{5mm} Taoyue Wang\su{1}\hspace{5mm} Xiaotian Li\su{1}\hspace{5mm}  Lijun Yin\su{1}}\\
    \vspace{2mm}
    {\large \hspace{6mm}\su{1}State University of New York at Binghamton} \hspace{8mm} 
    {\large \su{2}Rice University} \\
    {\tt\small \{zxiang4, twang61, xli210, lyin\}@binghamton.edu} \hspace{5mm} 
    {\tt\small hy48@rice.edu}
}
}

\maketitle

\begin{abstract}
Recent studies have focused on utilizing multi-modal data to develop robust models for facial Action Unit (AU) detection.
However, the heterogeneity of multi-modal data poses challenges in learning effective representations. 
One such challenge is extracting relevant features from multiple modalities using a single feature extractor. 
Moreover, previous studies have not fully explored the potential of multi-modal fusion strategies. 
In contrast to the extensive work on late fusion, there are limited investigations on early fusion for channel information exploration.
This paper presents a novel multi-modal reconstruction network, named \B{M}ultimodal \B{C}hannel-\B{M}ixing (\B{MCM}), as a pre-trained model to learn robust representation for facilitating multi-modal fusion.
The approach follows an early fusion setup, integrating a Channel-Mixing module, where two out of five channels are randomly dropped.
The dropped channels then are reconstructed from the remaining channels using masked autoencoder.
This module not only reduces channel redundancy, but also facilitates multi-modal learning and reconstruction capabilities, resulting in robust feature learning.
The encoder is fine-tuned on a downstream task of automatic facial action unit detection.
Pre-training experiments were conducted on BP4D+, followed by fine-tuning on BP4D and DISFA to assess the effectiveness and robustness of the proposed framework. 
The results demonstrate that our method meets and surpasses the performance of state-of-the-art baseline methods.
\end{abstract} 

\section{Introduction}
\label{sec:intro}

\begin{figure}[t]
  \centering
  \includegraphics[width=\linewidth]{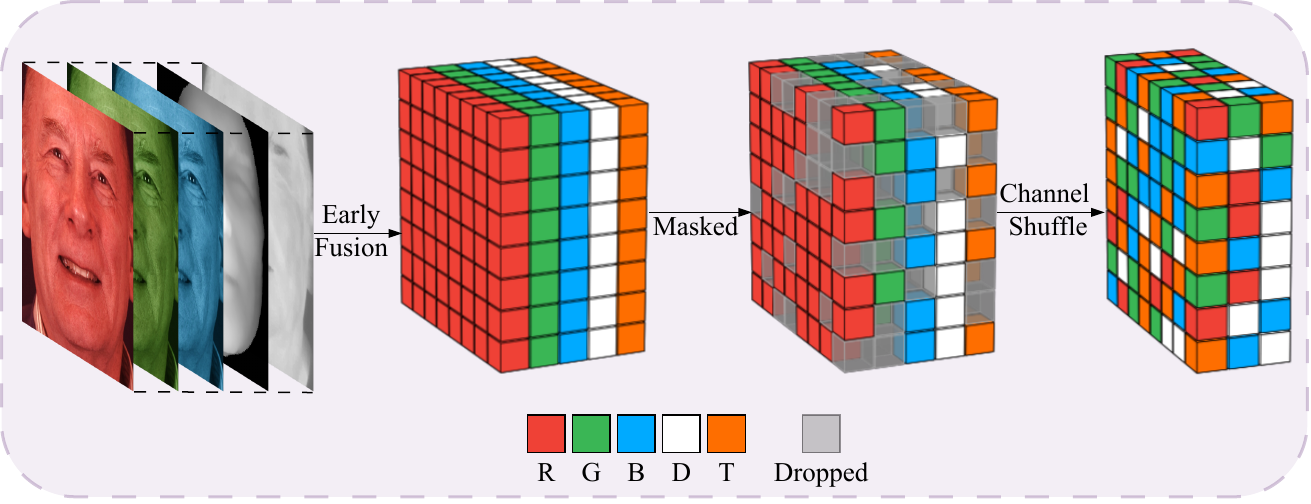}
  \caption{The Channel-Mixing module in our approach involves concatenating the channels of visual RGB, Depth, and Thermal modalities \IT{(RGBDT)}. 
  Subsequently, two channels are randomly dropped, and the three remaining channels in each patch are also randomly ordered. 
  Hence, our pre-training task is to reconstruct the missing channels from the remaining ones. 
  }
  \label{fig:channel_mixing}
  \vspace{-2mm}
\end{figure}

Facial Action Coding System (FACS)~\cite{Ekman1997} is designed to detect subtle facial changes by describing facial muscle movement, while the combination of basic AUs (Action Units) characterizes human expressions.
Automatic AUs detection has been studied for decades in human emotion analysis and human facial behavior analysis. 
AUs may occur individually or in combinations, hence AUs detection has been mainly treated as a multi-label classification problem, which is a challenge in computer vision.

In recent years, deep learning networks have shown popularity and advancement in many fields, e.g., computer vision, natural language processing, and autonomous driving.
Their successful application on large-scale datasets has also promoted a thriving research community on facial expression recognition and AU detection.
By using 2D images, several methods achieve state-of-the-art performance in AU detection\cite{eac,jaa,niu_2019,dsin,srerl,sev,au_transformer}.
However, these supervised approaches are very scarce for labeled data, and AU annotation is a highly labor-intensive and time-consuming process, thus researchers attempt to utilize unlabeled data through self-supervised learning\cite{LuTS20,tcae_2019}.
Moreover, as some subtle facial muscle movements are not easily observed in the images, multimodal approaches are necessary for efficient AU detection.
For instance, \textit{AU6} (cheek raiser) involves the deformation of \textit{Orbicularis oculi} and \textit{pars orbitalis} muscles, which shows subtle differences when observed in images but appears distinct geometric variations on the 3D mesh.
Similarly, human emotional variations can lead to changes in blood pressure, skin surface temperature, and heart rate, which are captured by physiological machines and thermal cameras.

Thanks to recently developed multi-modal databases\cite{wang2012,bp4d+,bp4d}, progress has been made by exploiting additional information in multi-modal learning on human emotion tasks.
For example, Li et al.~\cite{li2015,li2017} combined the 2D and 3D features  for facial expression recognition. 
Irani et al.~\cite{irani2015} utilized the visual, depth, and thermal modalities for pain study. 
In AU detection, Zhang et al.~\cite{mft} and Yang et al.~\cite{amf} present some fusion network with visual RGB, depth, or thermal images. 
They're all based on late or middle fusion, where each modality feature was extracted by a separate encoder, and fused into the latent space.
Late fusion methods have the advantage of dealing with missing modes, but they are often accompanied by an increase in network parameters while ignoring low-level modal interactions.
In contrast, early fusion, as the simplest example of fusion by concatenating individual modality features, hasn't been fully explored in recent multimodal learning studies and is typically used as a baseline method for comparison.
Recent works~\cite{depth_aware,single_stream_2020} on early fusion, have explored the utilization of visual RGB and depth modalities. 
However, in these approaches, the two modalities are simply concatenated on the channel dimension, which is a relatively under-explored.
To address these problems, we propose a multi-modal method named Multi-modal Channel-mixing Network (MCM).
This model not only learns early fusion features through an encoder, but also handle the downstream tasks even when some modalities are missing.
Firstly, the 5-channel input from the early fusion of RGB, Depth, and Thermal is processed through a patch-based Channel-Mixing (CM) module, resulting in a 3-channel output. 
Figure \ref{fig:channel_mixing} shows the overview structure of the channel mixing module.
Therefore, the channel mixing module ensures that the resulting 3-channel output is compatible with the backbone pre-trained on ImageNet~\cite{imagenet}. 
Additionally, this design allows for adaptation to self-supervised learning tasks such as image reconstruction.
Secondly, the mixed image is inputted into a ViT-based masked encoder, which extracts the fused latent feature. 
Subsequently, the latent features, along with mask tokens, are fed into three parallel ViT-based decoders to reconstruct the visual RGB, depth, and thermal images.
Similar to heavy spatial redundancy in an image, e.g., a missing patch can be recovered from neighboring patches~\cite{mae_2021}, channel redundancy also exists across in visual-depth-thermal data.
Therefore, the Channel-Mixing (CM) module not only reduces channel redundancy but also enhances multi-modal learning and reconstruction capabilities, leading to robust feature learning.

This architecture draws inspiration from the concept of masked autoencoder (MAE)~\cite{mae_2021}, expanding its scope from single modality to multiple modalities, from spatial masks to channel masks, and from object recognition to facial action unit detection.
Compared with related work, our contributions are as follows:
\begin{enumerate}

\item A multimodal channel-mixing (\IT{MCM}) scheme has been proposed for building a multi-modal pre-training framework. Such a novel scheme not only helps to learn a robust representation but also facilitates multi-modal fusion. 
\item
\hy{We have also proposed a unified framework for visual RGB, depth, and thermal modalities. It is more effective and efficient than the conventional multi-modal fusion strategies. Additionally, our proposed framework demonstrates good performance when some modalities are missing.}
\item
Extensive experiments have been conducted on three public datasets to evaluate the effectiveness of the proposed method. The results show that our model reaches and surpasses the state-of-the-art.
 
\end{enumerate}

\section{Related Work}

Automatic facial action recognition can be roughly categorized into single-modality-based methods and multi-modality-based approaches. 
The most commonly used modality is visual RGB image and a broad range of previous works ~\cite{Relatedworks_1, eac, Relatedworks_2,jaa,Relatedworks_4,Relatedworks_5, ksrl} has studied to use of a region of interest (ROI), image patches, facial landmarks, heat-map, attention mechanism, and other methods to localize detailed facial parts and learn discriminative feature representations. 
Considering the structural information and dependencies among different AUs on human faces, some recent works ~\cite{au_transformer, sev, mft} utilized the self-attention mechanism of transformer~\cite{transformer} to learn the semantic dependency among different AUs. 

Unstructured real-world data can inherently take many forms, also known as modalities, often including diverse representations of content. 
Multi-modal learning refers to an embodied learning situation that engages multiple sensory systems and action systems of the model. 
In the past decade, multi-modal learning has been studied on both facial expression recognition and action unit detection tasks. 
Li et al.~\cite{li2017} proposed a deep fusion network to learn the optimal combination weights of 2D and 3D facial representations for multi-modal 2D+3D FER. 
Irani et al.~\cite{irani2015} proposed a method to utilize RGB-Thermal-Depth images for pain estimation.
Li et al.~\cite{Relatedworks_9} explained that EEG signals show a strong correlation with facial actions and eye blinking of both posed and spontaneous expressions. 
They utilize the early fusion of EEG features and RGB features to boost both posed and authentic facial action detection. 
Liu et al.~\cite{temt} designed a crossmodal translation network encoding the RGB images to reconstruct thermal images, which is an end-to-end model for AU detection. 
Yang et al.~\cite{amf} proposed a model called AMF, which included a feature scoring module to select the most relevant feature representations from different modalities. 
Recently, the transformer-based~\cite{transformer} attention mechanism has been widely applied in multi-modal features fusion.
For example, MulT~\cite{mult} first attempted to fuse multi-modalities by transformer, which was applied to image, audio, and text.
TransFuser~\cite{transfuser} incorporates the global context of the 3D scene into the feature extraction layers of different modalities.
MFT ~\cite{mft} utilizes the RGB-Depth features fusion by multi-head fusion attention to fuse AU features of two modalities in a transformer.

All of these are late fusion or middle fusion works, where each modality has its encoder to encode the latent features, which are then fused at a deeper level.
Hence the parameter size is increased than the single encoder and the low-level interaction between the modalities is ignored.
Recently, the line between multimodal representation and fusion has been blurred for models such as deep neural networks where representation learning interacts with classification or regression objectives~\cite{multimodal}.
Hence, we refer to early fusion as the immediate integration of the data source or feature from the extractor, often simply by concatenating them into a single vector as the input to the learning model.
Recently, compared to the number of late fusion works, only a few works utilize the early fusion strategy, e.g., ~\cite{depth_aware, single_stream_2020,Relatedworks_9}.
One disadvantage of early fusion is the trained model usually can not handle the missing modality.
Although the conversion technique such as projection can be applied, the input size of the model is fixed.

Inspired by the success of the masked language model~\cite{bert,gpt,gpt_2,gpt_3} in NLP, masked image encoding methods of self-supervised learning have been improved in computer vision, often focusing on different pretext tasks for pre-training.
BEiT~\cite{beit_2022} design an autoencoder network similar to BERT~\cite{bert} with the ViT encoder and discrete image token. 
MAE~\cite{mae_2021} proposed another ViT backboned network that only encodes the visible patches and inserts masked tokens in decoding for reconstruction, which speeds up the training process.
Inspired by MAE, we propose a multimodal channel mixing (\textbf{MCM}) module to explore the effects of multimodal channel-based masking mechanism.
\B{Differ} from the previous late fusion works, our method only needs one encoder; 
\B{Differ} from the early fusion works, our downstream model can be fed with only one modality.

\begin{figure*}[!ht]
  \centering
  \includegraphics[width=0.95\textwidth]{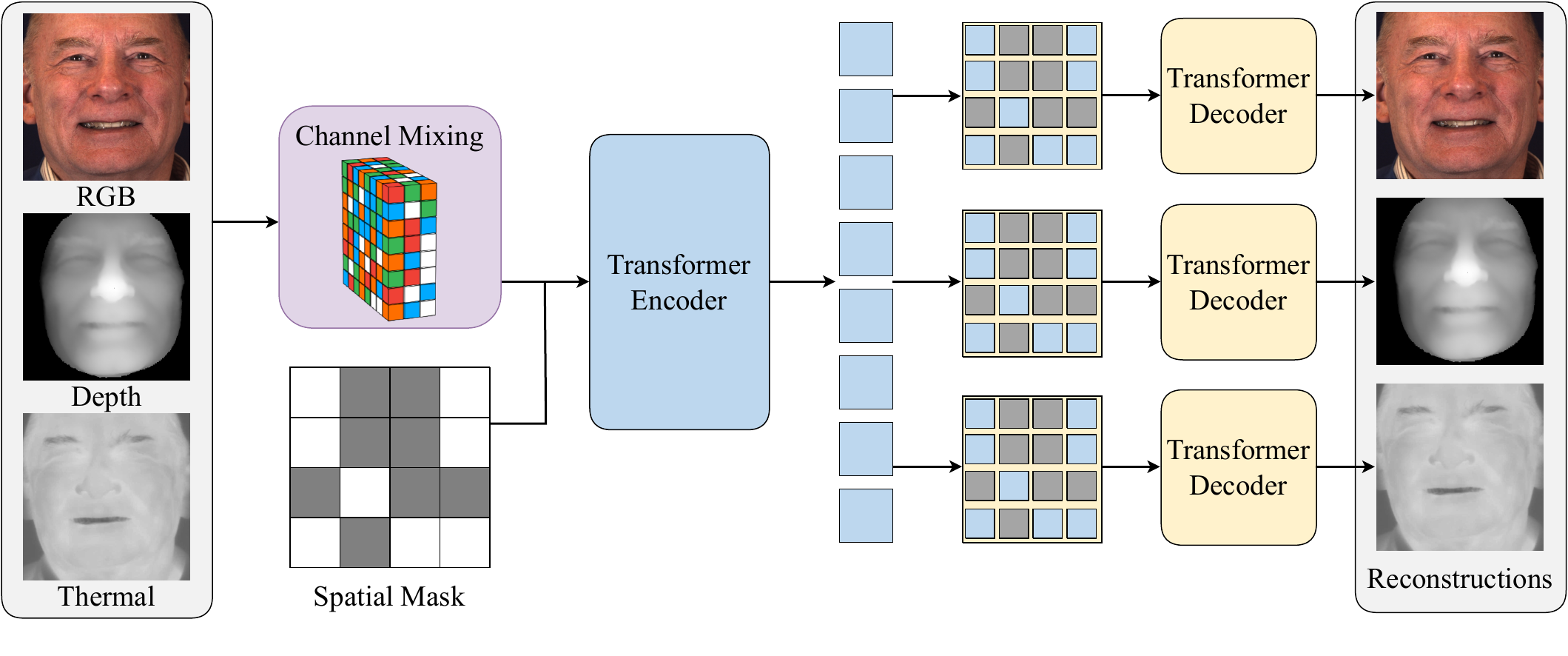}
  \caption{An overview of the proposed network, which is a multi-modal reconstruction network as a pre-train model for AU detection. 
  In this model, 3 modalities are fused with 5 channels and randomly dropped 2 channels, resulting in 3 channels. 
  Then a large subset of random patches is masked out, while the encoder is applied to the remaining visible patches. 
  Three subsets of masked tokens combined with the shared visible encoder patches are proceeded by three small decoders to reconstruct three modalities respectively. 
  Encoder and decoders are based on ViT and designed asymmetrical, i.e., 12, 8, 4, and 4 layers in the encoder, RGB decoder, Depth decoder, and Thermal decoder respectively.}
  \label{fig:architecture}
\end{figure*}

\section{Methodology}

In this section, we describe our method in detail, including Channel-Mixing Module, Masked Autoencoder, and AU detection network. The overall pipeline of our framework is illustrated in Figure~\ref{fig:architecture}.
Multimodalities are fed into the channel mixing module to get the 3-channels output. 
Following a spatial mask, the patches are encoded by a ViT encoder. 
The set of encoded patches and three sets of mask tokens are processed by three decoders that reconstruct the original RGB, Depth, and Thermal respectively.

\subsection{Channel-Mixing}

We have visual RGB image $I_{v}$ and its corresponding depth image $I_{d}$, and thermal image $I_{t}$, so the fused image is $I_{f} = I_{v} \oplus I_{d} \oplus I_{t}$, where $\oplus$ represents concatenation channels.
Then the fused image $i_f \in \mathbb{R}^{H\times W\times C} $ is split to patches $i_p \in \mathbb{R}^{L\times (p^2\cdot C)}$, where $(H, W)$ is its resolution, $C$ is the number of channels ($C=5$ in our task), $(P, P)$ is the resolution of each image patch, and $L = HW/P^2$ is the resulting number of patches~\cite{vit}.
Then each patch is randomly dropped with two channels and then shuffled in the channel, so the channel-mixing image set is $I_{mix} = \{i_{mix}\}$, where $i_{mix} \in \mathbb{R}^{L\times (p^2\cdot (C-2))}$.
Detail see in Figure~\ref{fig:channel_mixing}.

The proposed channel mixing module has the following advantages:
\begin{enumerate}
\item It downsamples the fused modality to 3 channels, which can take advantage of the large pre-trained parameters on Imagenet, allowing the AU detection model to potentially accept all modalities as input.
\item It increases the difficulty of the reconstruction task so that the network focuses on both spatial and channel information.
\item The encoder is guided by the reconstruction task to learn the most relevant latent feature for all three modalities, resulting in an effective and robust multimodal representation. 
\end{enumerate}

\subsection{Spatial Mask}

Patches are randomly removed based on a uniform distribution.
This approach has the potential to mask more patches near the center of the image, where the face is typically located.
With high masking rates, redundancy is largely eliminated, resulting in a task that is not easily solved by visible neighboring patches inference.
Since the following encoder only proceeds on unmasked patches, it speeds up the training process.
In addition, we assess the performance of our model using various masking rates. The details and results of these evaluations are presented in the supplementary materials.

\subsection{ViT Encoder}

Our encoder is a ViT~\cite{vit}, and only proceeds with the visible, unmasked patches. 
Unlike using learnable 1D position embedding in the standard ViT, we use the 2-D sine-cosine variant~\cite{mocov3_2021} for position embedding.
The embedding is then encoded by a stack of Transformer blocks following the design in~\cite{vit}.
Let $w_l$ be the encoded features at the l-th layer in transformer blocks, $W_0$ being the input layer. 
The features at the l-th layer are obtained by applying a transformer block defined as:
\begin{equation}\label{eq:t1}
    w_l^{\prime} = F_{LN}^l (w_{l-1} + F_{MSA}^l(w_{l-1}))
\end{equation}
\begin{equation}\label{eq:t2}
        w_l = F_{LN}^l (w_l^{\prime} + F_{FF}^l(w_l^{\prime}))
\end{equation}
where $F_{MSA}(\cdot)$ is the multi-head self-attention, $F_{LN}$ is Layernorm, and $F_{FF}$ is a fully connected feed-forward network.
We apply the ViT-base with 12 blocks and 512-dim as the encoder in our model.
Different from previous multi-modal AU detection works~\cite{amf,mft}, there's only one encoder for three modalities in the proposed network.

\subsection{ViT Decoder}

Since we have three modalities of reconstruction task, three decoders are involved in our network.
The inputs of three decoders $De=\{de_v,de_d,de_t\}$ are consisting of the shared encoded visible patches $e$, and three modalities mask tokens $\{m_v, m_d, m_t\}$, so that we have $de_v = \{e, m_v\}$, $de_d = \{e, m_d\}$, and $de_t = \{e, m_t\}$, where $v,d,t$ represent RGB, Depth, and Thermal.
Each mask token is a shared embedding vector that represents a missing patch to be predicted in each decoder. 
Following the mask token design in ~\cite{mae_2021}, where removing the mask token from the encoder can not only improve the accuracy but also speed up the training process.
The decoder position embeddings are also added to the input, otherwise, it would have no information about their location in the image.
The decoder has stacks of Transformer blocks, which are also defined as  Equation.\ref{eq:t1} and Equation.\ref{eq:t2}.
The last layer of a decoder is a linear projection whose number of output channels equals the number of pixel values in a patch.
The decoders are only used during pre-training to conduct the reconstruction tasks and only the encoder is used later for multi-modal representation learning on AU detection.

In general, the last several layers in an autoencoder are more specialized for reconstruction but are less relevant for recognition~\cite{mae_2021}.
The last layer of the encoder is adapted to the recognition task in fine-tuning, which means the decoder has less impact on improvement.
Inspired by the asymmetric design of MAE, we also design different decoders for image, depth, and thermal due to the different complexity of reconstruction tasks, i.e., 12 layers of blocks in the encoder, 8, 4, and 4 layers of blocks in the decoder for visual RGB, depth, and thermal respectively.

\subsection{Multi-modal Reconstruction}

Our network reconstructs the shared latent feature by predicting the pixels for each masked patch on three modalities.
Each element of the decoder output is a vector of pixel values representing a patch.
For each modality, the loss function computes the mean squared error (MSE) between the masked patches of the reconstructed and original image.
Define reconstruction losses $\mathcal{L}_{R}^{rgb}$ for RGB, $\mathcal{L}_{R}^{depth}$ for depth, and $\mathcal{L}_{R}^{thermal}$ for Thermal.
Then we combine these losses from three modalities by,
\begin{equation}
    \mathcal{L}_{R} = \mathcal{L}_{R}^{rgb} + \mathcal{L}_{R}^{depth} + \mathcal{L}_{R}^{thermal}
\end{equation}


\subsection{AU Detection}

We do supervised training to evaluate the representations from the encoder with end-to-end fine-tuning on AU detection.
Our model can accept mixed images as well as single modal images, where input $x \in \{ I_v, I_d, I_t, I_{mix} \}$.
Similar to the MAE~\cite{mae_2021} we use the average pooling instead of the class token.
As a multi-label task, AU detection faces the common problem of data unbalances.
Hence, we apply the weighted binary cross-entropy (BCE) loss functions. 
The $AU_k$ occurrence probability function $P(AU_k)$ and the correlate positive weight $p_k$ of weighted BCE are given as:

\begin{align}
        P(AU_k) &= \frac{\sum_{i=1}^{N} y_k^i}{\sum_{i=1}^{N} \sum_{k=1}^{C} y_{k}^i}\\
        p_k &= \frac{P(AU_k)}{\min\{P(AU_1), ..., P(AU_C)\}}
\end{align}

where $N$ is the total number of training set, $C$ is the number of AUs. 
Then the weighted BCE loss is defined:
\begin{align}
    \mathcal{L} =& - \frac{1}{N}\sum_{i=1}^{N} \sum_{k=1}^{C}(p_k\times y_k^i \times log(\widehat{y_k^i}) + \nonumber\\
       &\qquad{} (1-y_k^i) \times log(1-\widehat{y_k^i}))
\end{align}
where $y_k^i$ and $\widehat{y_k^i}$ represent the ground truth label and prediction for $AU_k$ respectively.

\section{Experiment}
Our pre-trained MCM is trained on the BP4D+~\cite{bp4d+}, where 300,000 random sample frames from all the frames are extracted (with/without AU notation).
The pretraining needs three modalities for input, that is RGB, Depth, and Thermal.
After the pretraining, we employ supervised training to finetune the encoder of AU detection on BP4D and DISFA~\cite{disfa}.
Through our experiments, we aim to address the research questions with respect to the following four aspects: 
\B{RQ1: }\IT{Can our proposed method outperform state-of-the-art baseline methods?} (section~\ref{single_result}.);
\B{RQ2: }\IT{Will multiple modalities fusion be more effective?} (section~\ref{mm_result}); 
\B{RQ3: }\IT{ What are the image reconstruction results across the database?};  (section~\ref{img_rec});
\B{RQ4: }\IT{How effective is MCM, and how efficient is it compared to traditional multimodal fusion methods?} (Section~\ref{effect_cm}).

\subsection{Data}

\noindent\B{BP4D~\cite{bp4d}}
contains 41 subjects (23 females and 18 males) captured in laboratory environments.
There are 8 tasks designed to elicit different spontaneous emotions.
Expert coders select the most expressive 20s of each video clip for AU coding, resulting in 140,000 labeled frames.
These labeled frames are used in AU detection fine-tuning.
In the AU detection experiments, following the work~\cite{eac}, we split them into subject-exclusive 3-fold with 12 AUs. 
Besides the RGB images, the database provides the corresponding 3D face meshes as well, generating the Depth images to use.
\B{BP4D+}~\cite{bp4d+}
consists of 140 subjects with a total of 1.5 M frames in the same laboratory environments.
In \B{pre-training}, we equally sample 300,000 frames from all 1.5 M frames by 10 activity categories.
Not only the RGB images, their corresponding Depth, and Thermal images are also used.
\B{DISFA~\cite{disfa}}
contains videos from the left view and right view of 27 subjects (12 females, 15 males). 
12 AUs are labeled with AU intensity from 0 to 5, resulting in around 130,000 AU-coded images. 
We choose 8 of 12 AUs from the left camera with AU intensities higher or equal to 2 as positive samples.
The recognition model trained on BP4D is then fine-tuned to the DISFA dataset, which is following the setting in ~\cite{eac, srerl}.
F1-score is reported based on subject-exclusive 3-fold cross-validation.


\begin{table*}[ht]
\caption{F1 scores in terms of 12 AUs \hy{are reported for the proposed method and the state-of-the-art methods }based on single modality on the BP4D. Bold numbers indicate the best performance in each modality.}
\small
\label{tab:table1}
\centering %
\renewcommand{\arraystretch}{1.0}
\begin{tabular}{l|c|*{12}{c}|c}
\hline
Methods     & Modal   & AU1       & AU2       & AU4       & AU6       & AU7       & AU10      & AU12      & AU14      & AU15      & AU17      & AU23      & AU24      & Avg\\
\hline \hline
EAC         & RGB     & 39.0      & 35.2      & 48.6      & 76.1      & 72.9      & 81.9      & 86.2      & 58.8      & 37.5      & 59.1      & 35.9      & 35.8      & 55.9\\
DSIN        & RGB     & 51.7      & 40.4      & 56.0      & 76.1      & 73.5      & 79.9      & 85.4      & 62.7      & 37.3      & 62.9      & 38.8      & 41.6      & 58.9\\
JAA-Net     & RGB     & 47.2      & 44.0      & 54.9      & 77.5      & 74.6      & 84.0      & 86.9      & 61.9      & 43.6      & 60.3      & 42.7      & 41.9      & 60.0\\
SRERL       & RGB     & 46.9      & 45.3      & 55.6      & 77.1      & 78.4      & 83.5      & 87.6      & 63.9      & 52.2      & 63.9      & 47.1      & 53.3      & 62.9\\ 
HMP-PS      & RGB     & 53.1      & 46.1      & 56.0      & 76.5      & 76.9      & 82.1      & 86.4      & 64.8      & 51.5      & 63.0      & 49.9      & 54.5      & 63.4\\
SEV-Net     & RGB     & \B{58.2}  & \B{50.4}  & 58.3      & \B{81.9}  & 73.9      & \B{87.8}  & 87.5      & 61.6      & 52.6      & 62.2      & 44.6      & 47.6      & 63.9\\
FAUT        & RGB     & 51.7      & 49.3      & \B{61.0}  & 77.8      & 79.5      & 82.9      & 86.3      & \B{67.6}  & 51.9      & 63.0      & 43.7      & 56.3      & 64.2\\
MFT         & RGB     & 49.6 	  & 48.1      & 59.9      & 78.4      & 78.0	  & 83.7  	  & 87.9      & 62.0	  & 55.3      & 61.8      & 50.9      & 54.9      & 64.2\\
KSRL        & RGB     & 53.3      & 47.4      & 56.2      & 79.4      & \B{80.7}  & 85.1      & 89.0      & 67.4      & 55.9      & 61.9      & 48.5      & 49.0      & 64.5\\
\B{MCM}     & RGB     & 54.4      & 48.5	  & 60.6	  & 79.1	  & 77.0	  & 84.0	  & \B{89.1}  & 61.7	  & \B{59.3}  & \B{64.7}  & \B{53.0}  & \B{60.5}  & \B{66.0}\\
\hline \hline
ResNet-18   & Depth   & 40.9	  & 40.7	  & 56.9      & 77.1      & 76.7      & 79.8      & 88.1      & 58.8      & 48.2      & 60.8      & 49.4      & 46.0      & 60.3\\
ResNet-50   & Depth   & \B{44.5}  & \B{41.3}  & \B{57.2}  & 76.5	  & 72.6	  & 80.9	  & 86.8	  & 54.3	  & 50.2      & 62.7      & 48.1	  & 44.5      & 60.0\\ 
MFT         & Depth   & 38.6	  & 37.3	  & 44.2	  & \B{84.2}  & \B{89.0}  & \B{89.7}  & \B{89.2}  & \B{79.8}  & 44.7	  & 46.0	  & \B{53.2}  & 37.0	  & 61.1\\
\B{MCM}     & Depth   & 44.3	  & 40.8	  & 51.3	  & 77.8	  & 70.9	  & 82.9	  & 87.3	  & 65.5	  & \B{50.8}  & \B{63.1}  & 48.1	  & \B{54.5}  & \B{61.4}\\ 
\hline
\end{tabular}
\end{table*}


\begin{table*}[ht]
\caption{F1 scores in terms of 8 AUs are reported for the proposed method and the state-of-the-art methods on the DISFA dataset. Bold numbers indicate the best performance}
\small
\label{tab:table2}
\centering %
\renewcommand{\arraystretch}{1.0}
\begin{tabular}{l|c|*{12}{c}|c}
\hline
Methods     & Modal   & AU1       & AU2       & AU4       & AU6       & AU9       & AU12      & AU25      & AU26      & Avg\\
\hline \hline
EAC         & RGB     & 41.5      & 26.4      & 66.4      & 50.7      & \B{80.5}  & \B{89.3}  & 88.9      & 15.6      & 48.5\\
DSIN        & RGB     & 42.4      & 39.0      & 68.4      & 28.6      & 46.8      & 70.8      & 90.4      & 42.2      & 53.6\\
JAA-Net     & RGB     & 43.7      & 46.2      & 56.0      & 41.4      & 44.7      & 69.6      & 88.3      & 58.4      & 56.0\\
SRERL       & RGB     & 45.7      & 47.8      & 59.6      & 47.1      & 45.6      & 73.5      & 84.3      & 43.6      & 55.9\\ 
SEV-Net     & RGB     & 55.3      & 53.1      & 61.5      & 53.6      & 38.2      & 71.6      & \B{95.7}  & 41.5      & 58.8\\
HMP-PS      & RGB     & 38.0      & 45.9      & 65.2      & 50.9      & 50.8      & 76.0      & 93.3      & 67.6      & 61.0\\
FAUT        & RGB     & 46.1      & 48.6      & \B{72.8}  & 56.7      & 50.0      & 72.1      & 90.8      & 55.4      & 61.5\\
KSRL        & RGB     & \B{60.4}  & \B{59.2}  & 67.5      & 52.7      & 51.5      & 76.1      & 91.3      & 57.7      & \B{64.5} \\
\B{MCM}     & RGB     & 49.6      & 44.1      & 67.2      & \B{65.5}  & 49.0      & 81.5      & 85.9      & \B{71.8}  & 64.3\\
\hline
\end{tabular}
\end{table*}

\begin{figure}[h]
  \centering
  \includegraphics[width=0.95\linewidth]{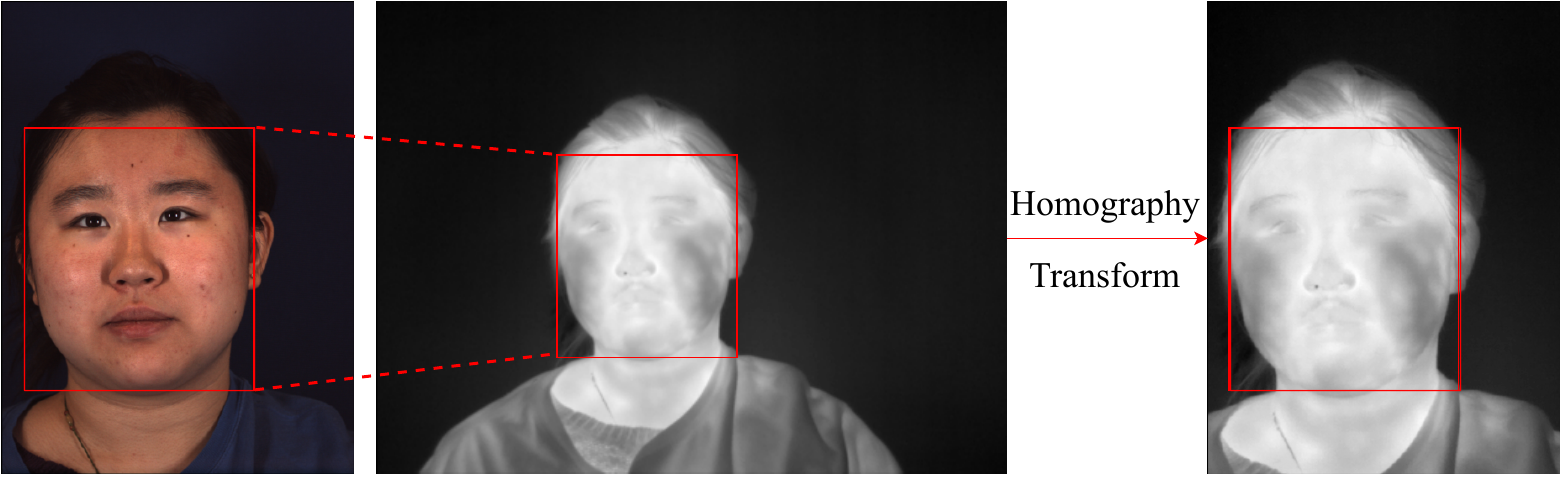}
  \caption{An example of Homography Transform on Thermal images based on their bounding boxes.}
  \label{fig:homo}
\end{figure}

\subsection{Preprocessing}
In our experiments, we use three modalities in BP4D+, 2D RGB, Depth, and Thermal images. 
The face regions of RGB images are detected and cropped by the publicly available library OpenCV.
A raw 3D face model contains a 30k - 50k number of vertices in BP4D databases, including the face, hair, and neck areas of a subject.
We first project the 3D mesh into depth images matching the visual image and then cropped the face area with the corresponding visual bounding box.
Thermal images are normalized based on the original thermal temperature data. 
Since the RGB camera and Thermal camera are positioned differently, resulting in varying images, we apply Homography Transform to align the Thermal images with their corresponding RGB images, which is shown in Figure.~\ref{fig:homo}.
For each subject, only one bounding box is necessary for RGB and one for Thermal images. 

\subsection{Implementation details}
\B{Pre-training.}
Following the standard ViT architecture, we use ViT-base as the encoder and decoders, where 12 layers are in the encoder, and 8, 4, and 4 layers in decoders for RGB, Depth, and Thermal, respectively. 
The imagenet pre-training weights from~\cite{mae_2021} are used to our encoder.
Optimizer Adamw~\cite{adamw_2019} with learning rate 2.e-4, weight decay 0.05, and momentum (0.9, 0.95).
We train 100 epochs with batch size 64 where the warmup epoch~\cite{warmup_2017} is set with 40 and then perform cosine decay~\cite{sgdr_2016} with minimal lr by 1.e-6.
\noindent\B{Fine-tuning}
The drop path~\cite{drop_path_2016} for ViT to 0.1 and layer-wise lr decay~\cite{beit_2022} is 0.75.
We set the batch size to 64, train 5 epochs with no warmup and cosine decay in fine-tuning, and the reduction factor of lr is 10 at the 3rd epoch.

We implement our method with Pytorch-1.11.0 and cuda-11.3 on the GPU of NVIDIA RTX 3090. 
To evaluate the performance of AU detection accuracy, we use the F1-score for a comparative study with the state-of-the-art.


\begin{table*}[h]
\caption{F1 scores in terms of 12 AUs are reported for the proposed method and the state-of-the-art methods based on RGB+Depth modalities on BP4D. Bold numbers indicate the best performance}
\small
\label{tab:table3}
\centering %
\renewcommand{\arraystretch}{1.0}
\begin{tabular}{l|c|*{12}{c}|c}
\hline 
Methods         & Modal       & AU1       & AU2       & AU4       & AU6       & AU7       & AU10      & AU12      & AU14      & AU15      & AU17      & AU23      & AU24      & Avg\\
\hline \hline
MTUT            & RGBD        & 51.3      & 50.2      & \B{62.2}  & 77.2      & 71.7      & 83.8      & 88.2      & 61.4      & 54.3      & 57.9      & 45.8      & 42.2      & 62.2\\ 
TEMT-Net        & RGBD        & 53.7      & 47.1      & 60.5      & 77.6      & 75.6      & 84.8      & 87.4      & 67.0      & 57.2      & 61.3      & 44.7      & 41.6      & 63.2\\ 
AMF             & RGBD        & 55.1      & \B{58.3}  & 62.0      & \B{82.5}  & 75.6      & \B{87.2}  & \B{89.6}  & 60.9      & \B{59.1}  & 62.4      & 45.0      & 52.0      & 65.8\\ 
MFT             & RGBD        & 51.6	  & 49.2	  & 57.6      & 78.8	  & 77.5      & 84.4	  & 87.9	  & 65.0	  & 56.5	  & 64.3      & 49.8	  & 55.1      & 64.8\\
\B{MCM}         & RGBD        & \B{56.5}  & 47.4	  & \B{62.2}  & 79.9	  & \B{78.2}  & 84.3	  & 87.9	  & \B{67.8}  & 55.8	  & \B{64.8}  & \B{52.3}  & \B{57.2}  & \B{66.2}\\ 
\hline
\end{tabular}
\end{table*}

\begin{figure*}[h]
     \centering
     \begin{subfigure}[b]{0.14\textwidth}
         \centering
         \includegraphics[width=\textwidth]{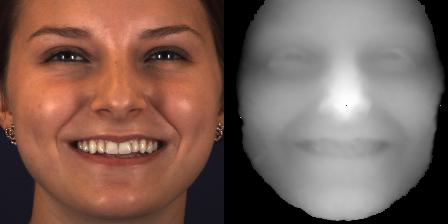}
     \end{subfigure}
     \hfill
     \begin{subfigure}[b]{0.28\textwidth}
         \centering
         \includegraphics[width=\textwidth]{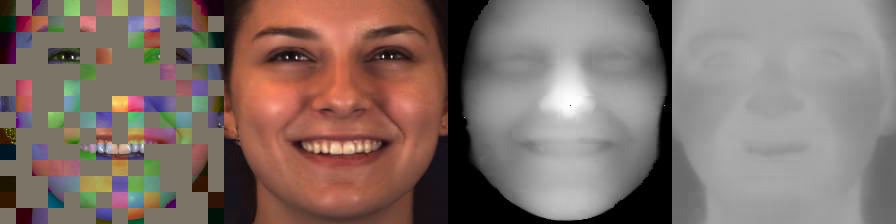}
     \end{subfigure}
     \hfill
     \begin{subfigure}[b]{0.28\textwidth}
         \centering
         \includegraphics[width=\textwidth]{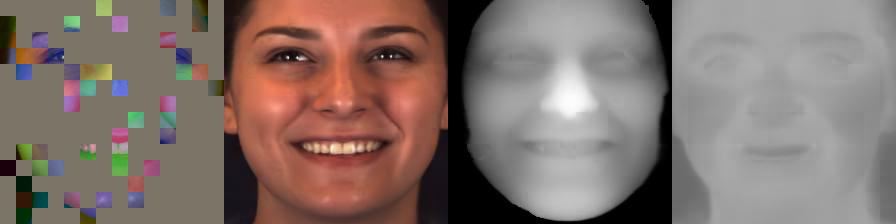}
     \end{subfigure}
     \hfill
     \begin{subfigure}[b]{0.28\textwidth}
         \centering
         \includegraphics[width=\textwidth]{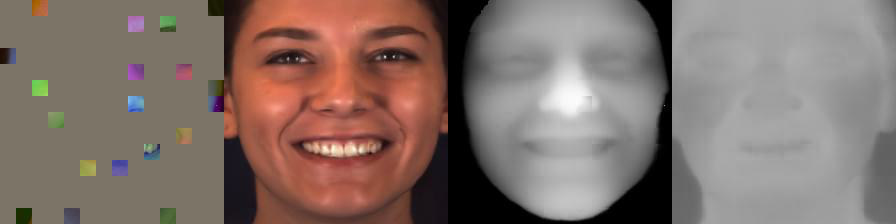}
     \end{subfigure}
     
      \begin{subfigure}[b]{0.14\textwidth}
         \centering
         \includegraphics[width=\textwidth]{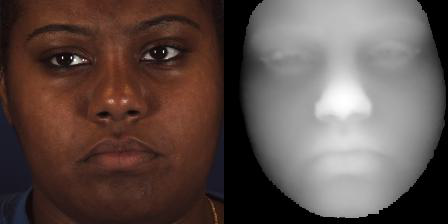}
     \end{subfigure}
     \hfill
     \begin{subfigure}[b]{0.28\textwidth}
         \centering
         \includegraphics[width=\textwidth]{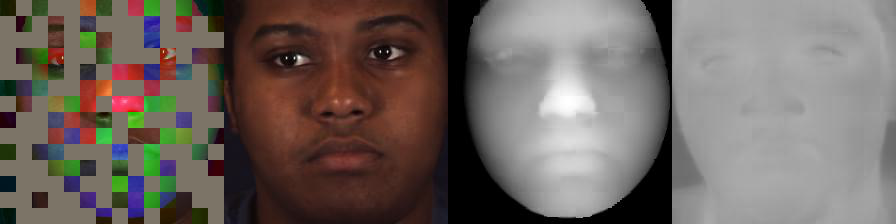}
     \end{subfigure}
     \hfill
     \begin{subfigure}[b]{0.28\textwidth}
         \centering
         \includegraphics[width=\textwidth]{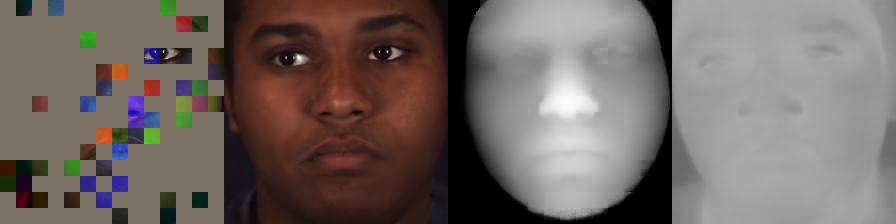}
     \end{subfigure}
     \hfill
     \begin{subfigure}[b]{0.28\textwidth}
         \centering
         \includegraphics[width=\textwidth]{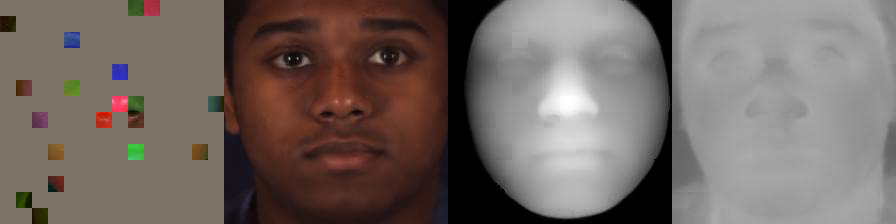}
     \end{subfigure}

     \begin{subfigure}[b]{0.14\textwidth}
         \centering
         \includegraphics[width=\textwidth]{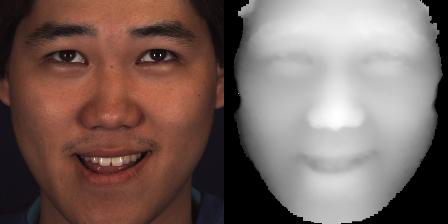}
     \end{subfigure}
     \hfill
     \begin{subfigure}[b]{0.28\textwidth}
         \centering
         \includegraphics[width=\textwidth]{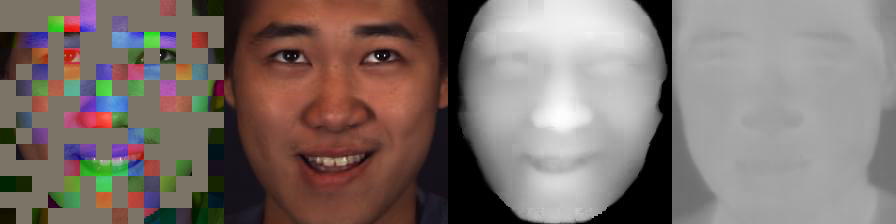}
     \end{subfigure}
     \hfill
     \begin{subfigure}[b]{0.28\textwidth}
         \centering
         \includegraphics[width=\textwidth]{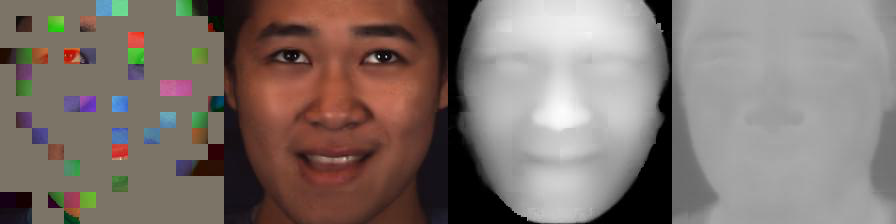}
     \end{subfigure}
     \hfill
     \begin{subfigure}[b]{0.28\textwidth}
         \centering
         \includegraphics[width=\textwidth]{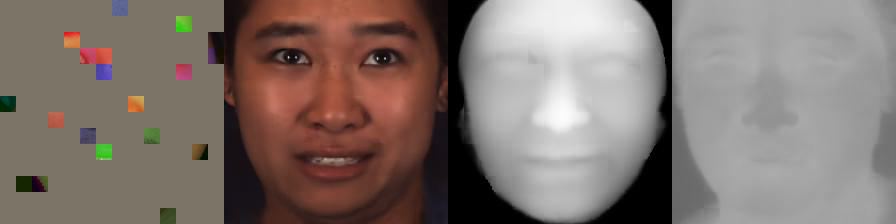}
     \end{subfigure}
     
     \begin{subfigure}[b]{0.14\textwidth}
         \centering
         \includegraphics[width=\textwidth]{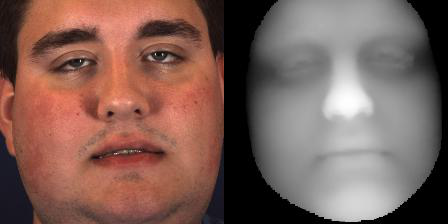}
         \caption{origin}
     \end{subfigure}
     \hfill
     \begin{subfigure}[b]{0.28\textwidth}
         \centering
         \includegraphics[width=\textwidth]{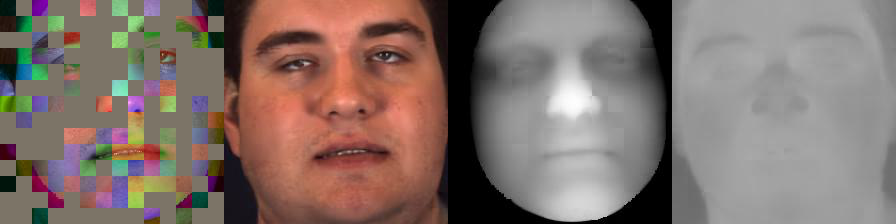}
         \caption{mask: 50\%}
     \end{subfigure}
     \hfill
     \begin{subfigure}[b]{0.28\textwidth}
         \centering
         \includegraphics[width=\textwidth]{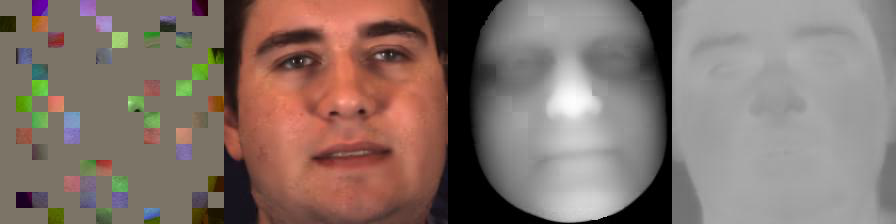}
         \caption{mask: 75\%}
     \end{subfigure}
     \hfill
     \begin{subfigure}[b]{0.28\textwidth}
         \centering
         \includegraphics[width=\textwidth]{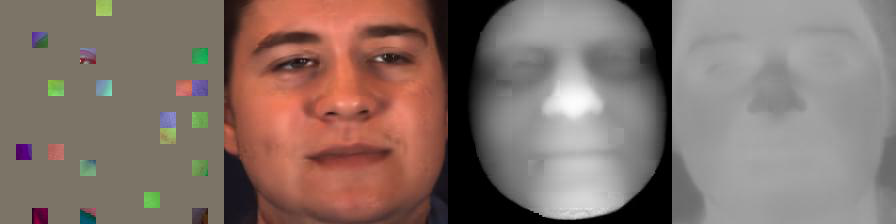}
         \caption{mask: 90\%}
     \end{subfigure}
\caption{
Visualization of the reconstructed images on BP4D with different mask ratios is performed using a reconstruction model trained on BP4D+. 
Images in (b, c, and d) represent the masked channel-mixing image, reconstructed RGB, reconstructed Depth, and reconstructed Thermal, respectively.
}
\label{fig:mask_samples_bp4d}
\end{figure*}

\subsection{Comparison with related methods}

\subsubsection{Single-modal results (\B{RQ1})}\label{single_result}

\noindent\B{RGB.} Our approach allows the use of a single modality in AU detection. 
Hence, we first compare our method to the state-of-the-art algorithms on an RGB modality in terms of the F1-score.
Methods included on BP4D are EAC\cite{eac}, DSIN\cite{dsin}, JAA-Net\cite{jaa}, SRERL\cite{srerl}, 
HMP-PS~\cite{hmpps_2021}, SEV-Net\cite{sev}, FAUT~\cite{au_transformer}, MFT~\cite{mft}, and KSRL~\cite{ksrl}. 
The top section in Table~\ref{tab:table1} shows the results of visual RGB images on BP4D, where our method achieves the best performance.
Specifically, our method outperforms the SOTA methods in recognizing \textit{AU4}, \textit{AU12}, \textit{AU15}, \textit{AU17}, \textit{AU23}, \textit{AU24}.
 
\noindent\B{Depth.} To evaluate our method on Depth modality, we compare with~\cite{mft} and Resnet-18, Resnet-50, as the baseline functions, where our method outperforms the state of the art. 
Since depth has only one channel, it's duplicated into 3 channels as the input when using Depth only.
The result is reported in the bottom section of Table.\ref{tab:table1}.

Table~\ref{tab:table2} provides the results on the DISFA dataset. 
The performance is evaluated on 8 action units, in line with prior works~\cite{eac,dsin,jaa,srerl,sev,hmpps_2021,au_transformer,ksrl}.
For DIFSA, following the protocol in ~\cite{eac, srerl}, we do not train the model from scratch, but fine-tune the model from training on BP4D. 
Our method achieves the state-of-the-art where only slightly less than KSRL~\cite{ksrl}.

\subsubsection{Multi-modal results (\B{RQ2})}\label{mm_result}

Since our method also accepts a fused channel-mixing feature as the input, we compare our method with the state-of-the-art multi-modal methods, i.e., MTUT~\cite{mtut}, TEMT-Net~\cite{temt}, AMF~\cite{amf}, and MFT~\cite{mft}.
The F1-score of MTUT and TEMT-Net are reported in the work~\cite{amf}, where the author implemented them with the fusion of RGB and Depth.
The results on BP4D are shown in Table~\ref{tab:table3}. 
Our method outperforms all the related methods and achieves the highest F1-score at 66.2\%.
In terms of 12 AUs, our method performs best in 7 AUs: \textit{AU1},\IT{AU4}, \textit{AU7}, \textit{AU14}, \textit{AU17}, \textit{AU23} and \textit{AU24}.
The fusion feature surpasses the performance of all single-modality methods, which validates the effectiveness of utilizing multiple modalities.


\begin{table*}[t]
\caption{Ablation study of the effectiveness of proposed Channel-Mixing module on BP4D database. 
IM: input modality in detection; CM: Channel-Mixing modalities; 
V: Visual RGB; D: Depth; T: Thermal. 
F1 scores in terms of 12 AUs are reported for different variants on the BP4D. 
A noticeable improvement is shown by using CM in both two stages of Pretrain and Finetune.}
\small
\label{tab:table4}
\centering %
\renewcommand{\arraystretch}{1.0}
\begin{tabular}{c|c|c|*{12}{c}|c}
\hline
Stage           & IM        & CM        & AU1   & AU2   & AU4   & AU6   & AU7   & AU10   & AU12   & AU14   & AU15   & AU17   & AU23   & AU24   & Avg\\
\hline \hline
\MURot{6}{Pretrain}& V      & \xmark    & 53.1  & 45.4  & 56.3	& 80.2	& 74.8	& 84.1	& 88.5	& 67.1	& 52.6	& 67.2	& 53.4	& 57.8	& 65.0\\ 
                & V         & VD        & 53.4  & 48.5  & 57.8  & 79.8  & 76.7  & 83.1  & 88.5  & 64.2  & 54.9  & 64.5  & 53.5  & 61.2  & 65.5\\
                & V         & VDT       & 54.4  & 48.5	& 60.6	& 79.1	& 77.0	& 84.0	& 89.1  & 61.7	& 59.3  & 64.7  & 53.0  & 60.5  & \B{66.0}\\\cline{2-16}
                & D         & \xmark    & 44.5	& 41.4	& 51.0	& 77.5	& 69.8	& 82.1	& 87.2	& 62.1	& 48.3	& 65.7	& 48.7	& 53.1	& 61.0	   \\
                & D         & VD        & 44.3	& 40.8	& 51.3	& 77.8	& 70.9	& 82.9	& 87.3	& 65.5	& 50.8	& 63.1	& 48.1	& 54.5	& \B{61.4}\\
                & D         & VDT       & 44.0	& 39.2	& 50.4	& 77.0	& 71.1	& 82.9	& 87.6	& 64.4	& 52.4	& 62.6	& 48.6	& 55.6	& 61.3\\\cline{2-16}
                & VD        & \xmark    & 54.6	& 49.1	& 58.4	& 79.7	& 76.0	& 83.7	& 88.6	& 61.9	& 55.6	& 64.6	& 52.6	& 60.3	& 65.4 \\ 
                & VD        & VD        & 56.5  & 47.4	& 62.2  & 79.9	& 78.2  & 84.3	& 87.9	& 67.8  & 55.8	& 64.8  & 52.3  & 57.2  & \B{66.2}\\
                & VD        & VDT       & 56.9  & 47.9  & 59.4  & 79.6  & 76.1  & 83.4  & 89.2  & 62.0  & 59.6  & 63.8  & 52.8  & 59.2  & 65.8\\ 
\hline
\hline
\MURot{4}{Finetune}& I     & \xmark    & 43.9	& 42.9	& 51.1	& 78.9  & 77.1  & 83.2  & 87.6	& 69.6  & 51.9	& 60.7	& 46.2	& 58.8  & 62.7\\ 
                & D     & \xmark    & 40.6	& 36.7	& 50.8	& 78.8	& 72.1	& 81.8	& 87.8	& 68.1  & 50.0	& 61.7  & 47.0	& 55.3  & 60.9\\
                & VD      & \xmark    & 44.0	& 42.7	& 48.5	& 79.0	& 76.7	& 83.2	& 87.8	& 69.0	& 52.3	& 63.1	& 48.8	& 54.7	& 62.5 \\ 
                & VD      & \cmark    & 47.9	& 45.8	& 48.9	& 78.9	& 77.6	& 84.2	& 88.0	& 68.4	& 52.0	& 61.6	& 46.8	& 55.9	& \B{63.0} \\ 
\hline
\end{tabular}
\end{table*}

\begin{figure}[ht]
 \centering
    \begin{subfigure}[b]{\linewidth}
         \centering
         \includegraphics[width=\textwidth]{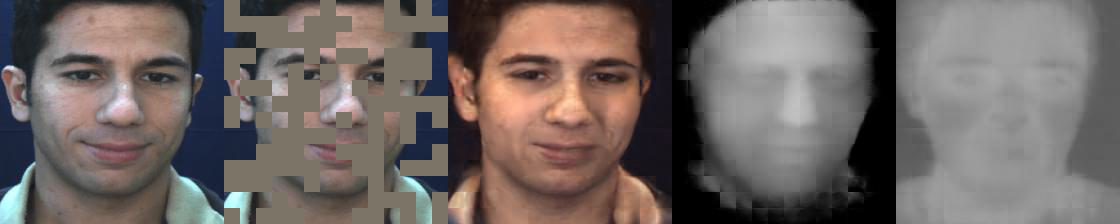}
     \end{subfigure}
     
     \begin{subfigure}[b]{\linewidth}
         \centering
         \includegraphics[width=\textwidth]{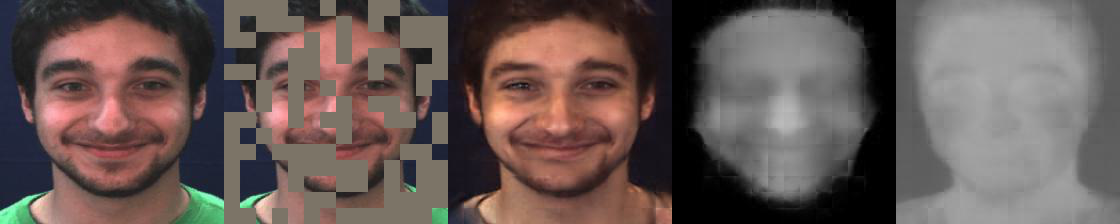}
     \end{subfigure}

      \begin{subfigure}[b]{\linewidth}
         \centering
         \includegraphics[width=\textwidth]{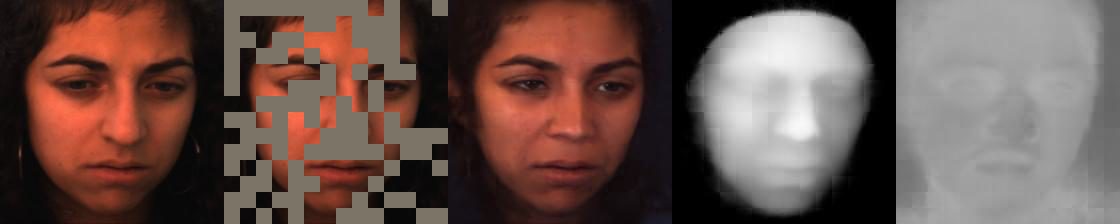}
     \end{subfigure}
     
      \begin{subfigure}[b]{\linewidth}
         \centering
         \includegraphics[width=\textwidth]{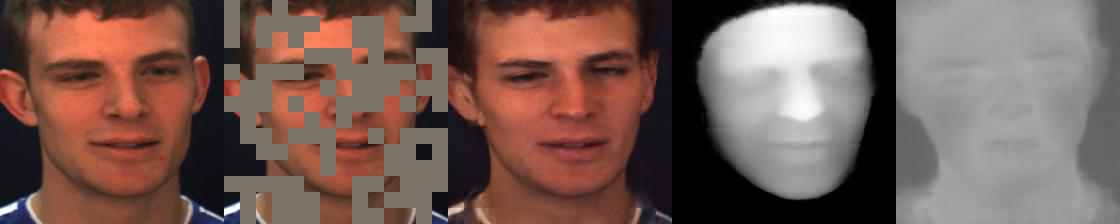}
     \end{subfigure}
     
  \caption{
  \hy{Visualization of the reconstructions on DISFA. The examples are reconstructed from the DISFA dataset using a pre-trained model on the BP4D+ and a mask ratio of 50\%.}}
  \label{fig:mask_samples_disfa}
  \vspace{-2mm}
\end{figure}

\subsection{Ablation Studies}

\subsubsection{Image reconstructions (\B{RQ3})}\label{img_rec}
Figure~\ref{fig:mask_samples_bp4d} shows reconstructions result on BP4D by different mask ratios, i.e., 50\%, 75\% and 90\%.
Inputs with RGB and Depth through the channel-mixing resulting the final reconstruction result in RGB, Depth, and Thermal.
The reconstructions are generated by the pre-trained model with 50\% masks.
As BP4D+ was using similar recording cameras to BP4D, our proposed method works well for the multi-modalities reconstruction task.
Although all the identities in BP4D are unseen to the model trained in BP4D+, it successfully reconstructs the facial images of individuals using only a few visible patches.
As the mask ratio increases, the quality of the reconstructed images tends to decrease, particularly with regards to preserving identity and expression. 
Higher mask ratios introduce more information loss, leading to more significant changes in the reconstructed images.

In the case of DISFA, where only RGB modality is available, our pre-training process utilizes only RGB images as the input.
This also demonstrates the flexibility and adaptability of our method in handling different modalities across different datasets.
The result is shown in Figure~\ref{fig:mask_samples_disfa}.
The RGB image reconstruction on DISFA is not as good as on BP4D, which is caused by the following reason.
The image color in DISFA is biased, i.e., the distribution of the image ranges from warm to cool colors, which is much more deviated than BP4D+.
See the enormous difference in the ground truth color between rows 1 and 3 in Figure~\ref{fig:mask_samples_disfa}.
In such a case, the reconstruction model learned from BP4D+ is quite difficult to match all the colors on DISFA.
The reconstructed images are quite warm, so the color from some examples is unmatched.
However, the semantic information of the face, e.g., eyes, nose, and mouth, is well reconstructed.
Meanwhile, due to the missing Depth modality, its Depth reconstruction result is also worse than BP4D.
Considering there's no same identity person across BP4D, BP4D+, and DISFA, our framework has a good quality reconstruction across the databases.

\subsubsection{Effectiveness and Efficiency of CM (\B{RQ4})} \label{effect_cm}

\noindent\B{Effectiveness.} We conducted comparison experiments on the BP4D dataset to carefully analyze the contribution of our proposed Channel-Mixing (CM) module, which was utilized in both the pre-training and fine-tuning stages.
Table~\ref{tab:table4} shows the performance of various comparisons in terms of the F1-score of 12 AUs.
We first examine the CM module in the pre-training stage, where we report the detection result of all modalities fine-tuning on different pre-trained multi-modal models.
MAE without the CM module is used as the baseline function.
Applying the proposed channel-mixing achieves a noticeable increase at 1.5\%, 0.7\%, and 1.2\% in RGB, Depth, and RGBD respectively, leading to an improvement over the SOTA.
Then a pre-trained ViT is used to evaluate the CM module in the fine-tuning stage.
See the results in the Finetune section of Table~\ref{tab:table4}.
A 1x1 convolutional layer is used to reduce the 4-channel RGBD to 3 channels for embedding.
Performance in RGB achieves the best by this baseline model, which is better than multimodal early fusion.
However, by directly utilizing the CM module, the RGBD early fusion outperforms the RGB modality.
Comparing the two stages, MAE in pre-training improves more clearly.
In the pre-training, channel mixed features are reconstructed through the encoder and decoder, effectively utilizing existing channel information to generate the missing channels.
However, the mixing feature is fed directly to a ViT for AU pattern learning in fine-tuning.

\noindent\B{Efficiency}. The total parameter count of the pre-trained model is 125M, with 85M parameters located in the encoder. In comparison to the traditional three-encoder late fusion method, our proposed approach efficiently reduces approximately 57\% (170M) of trainable parameters.

\section{Conclusion}
This paper has proposed a novel multi-modal learning approach that utilizes visual RGB, depth, and thermal modalities to train a reconstruction network on pre-training.
We have designed a channel-mixing module that empowers our method to process either a single modality or multi-modalities using a single-stream encoder.
The proposed method is evaluated on three widely used facial behavior databases, BP4D, BP4D+, and DISFA, where the AU detection performance has surpassed or reached the state-of-the-art.
Additionally, the experiments demonstrate the effectiveness of multi-modal fusion, which outperforms single-modality approaches. 
Ablation experiments further validate the efficiency and robustness of the proposed channel mixing module.

{\small
\bibliographystyle{ieee_fullname}
\bibliography{egbib}
}

\end{document}


\title{Multimodal Channel-Mixing: Channel and Spatial Masked AutoEncoder on Facial Action Unit Detection Supplementary Material}

\maketitle


\section{Mask ratio and Training epoch in pre-training}

\begin{figure}[ht]
  \centering
    \begin{subfigure}[b]{\linewidth}
         \centering
         \includegraphics[width=\linewidth]{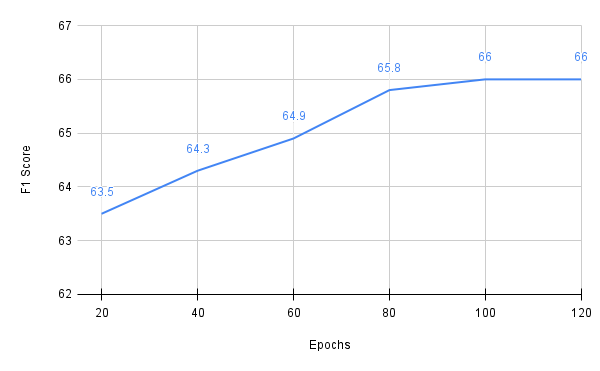}
         \caption{Fine-tune performance on various pre-training epochs, where a noticeable improvement along epochs.}
         \label{fig:epoch}
     \end{subfigure}
    \vspace{1.5mm}
    \begin{subfigure}[b]{\linewidth}
         \centering
         \includegraphics[width=\linewidth]{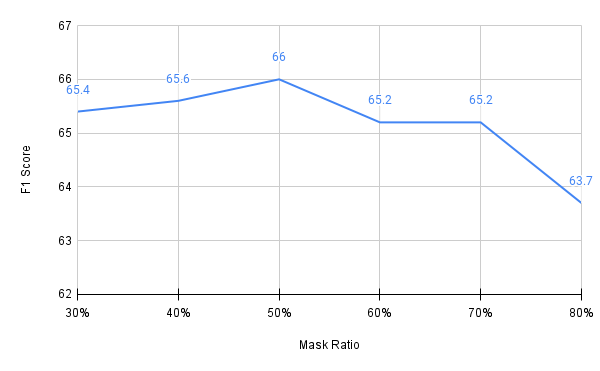}
         \caption{Fine-tune performance on various mask ratios of the pre-training model.}
         \label{fig:mask_ratio}
    \end{subfigure}
    
  \caption{ (a) F1 scores are reported by the model trained on BP4D+ with 50\% mask. (b) F1 scores in fusion are reported by the model trained on BP4D+ with epoch-100.}
  \label{fig:epoch_mask_ratio}
\end{figure}

First, the ablation study of pre-training epoche is based on a 50\% mask, where the result of RGB modality is shown in Figure~\ref{fig:epoch} in terms of average F1 scores.
Performance steadily improved during the training process and reached its optimum at the 100 epoch.
Our mask ratio experiments thus far are based on the pre-training model on epoch-100. 

Then we evaluate the masking ratio influence on downstream AU detection performance, see in Figure~\ref{fig:mask_ratio}.
Performance is improved as the mask ratio is increased, while the optimal ratio is 50\%.
When the ratio is greater than 50\%, the f1 score is obviously decreased, because much more detailed information is missed.
Our following reconstruction visualization is based on the model training on BP4D+ with 100-epoch and 50\% mask ratio.

\section{More Visualization Examples}

\begin{figure*}[th]
     \centering
     \begin{subfigure}[b]{0.14\textwidth}
         \centering
         \includegraphics[width=\textwidth]{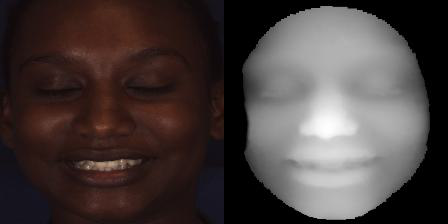}
     \end{subfigure}
     \hfill
     \begin{subfigure}[b]{0.28\textwidth}
         \centering
         \includegraphics[width=\textwidth]{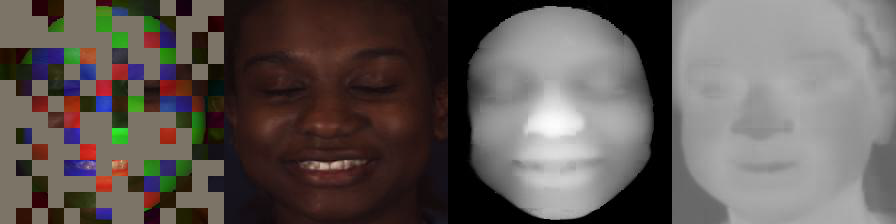}
     \end{subfigure}
     \hfill
     \begin{subfigure}[b]{0.28\textwidth}
         \centering
         \includegraphics[width=\textwidth]{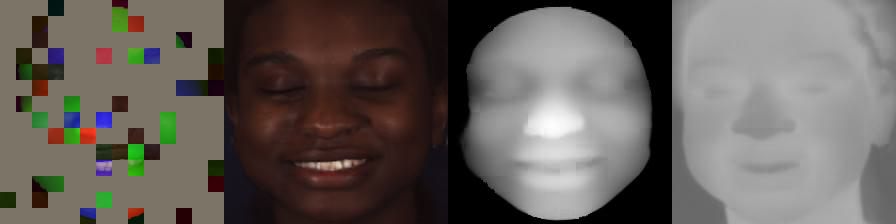}
     \end{subfigure}
     \hfill
     \begin{subfigure}[b]{0.28\textwidth}
         \centering
         \includegraphics[width=\textwidth]{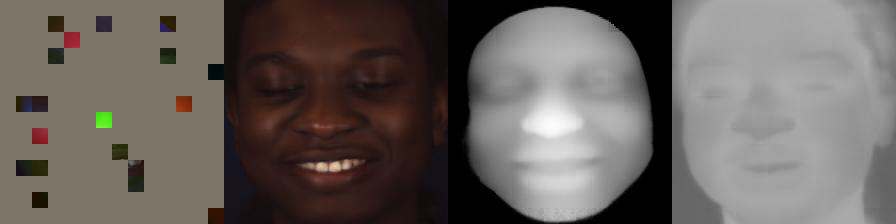}
     \end{subfigure}
     
      \begin{subfigure}[b]{0.14\textwidth}
         \centering
         \includegraphics[width=\textwidth]{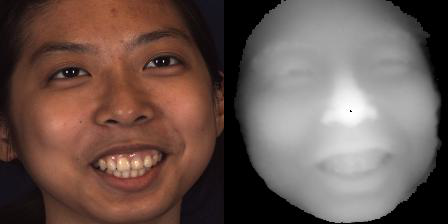}
     \end{subfigure}
     \hfill
     \begin{subfigure}[b]{0.28\textwidth}
         \centering
         \includegraphics[width=\textwidth]{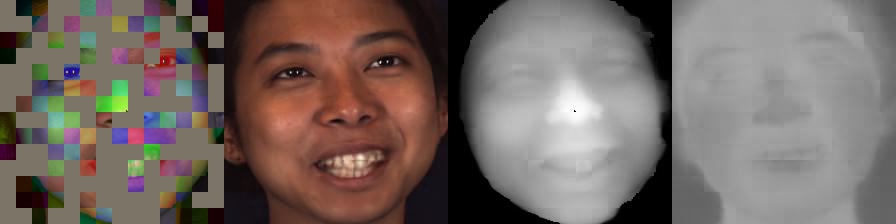}
     \end{subfigure}
     \hfill
     \begin{subfigure}[b]{0.28\textwidth}
         \centering
         \includegraphics[width=\textwidth]{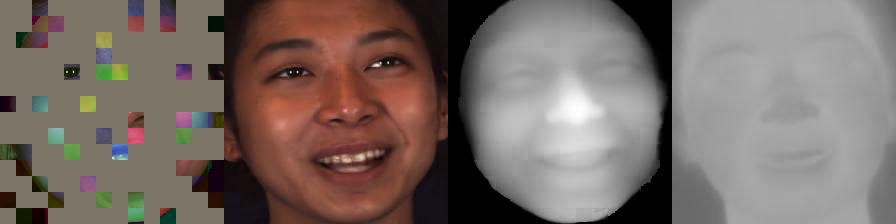}
     \end{subfigure}
     \hfill
     \begin{subfigure}[b]{0.28\textwidth}
         \centering
         \includegraphics[width=\textwidth]{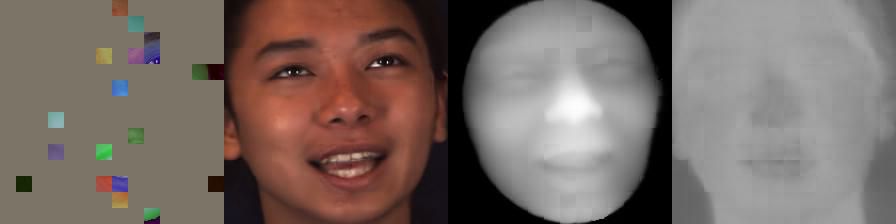}
     \end{subfigure}

     \begin{subfigure}[b]{0.14\textwidth}
         \centering
         \includegraphics[width=\textwidth]{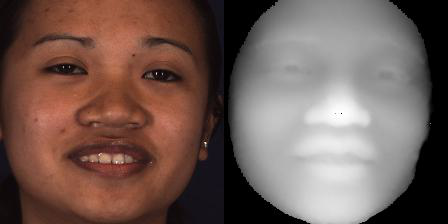}
     \end{subfigure}
     \hfill
     \begin{subfigure}[b]{0.28\textwidth}
         \centering
         \includegraphics[width=\textwidth]{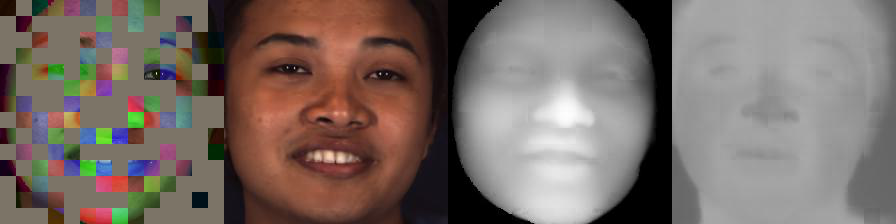}
     \end{subfigure}
     \hfill
     \begin{subfigure}[b]{0.28\textwidth}
         \centering
         \includegraphics[width=\textwidth]{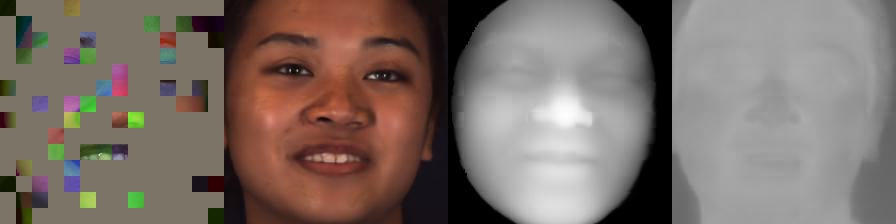}
     \end{subfigure}
     \hfill
     \begin{subfigure}[b]{0.28\textwidth}
         \centering
         \includegraphics[width=\textwidth]{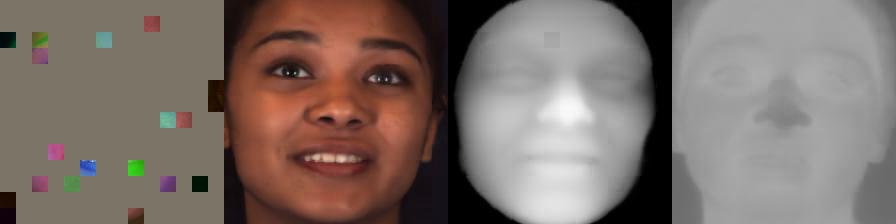}
     \end{subfigure}
     
     \begin{subfigure}[b]{0.14\textwidth}
         \centering
         \includegraphics[width=\textwidth]{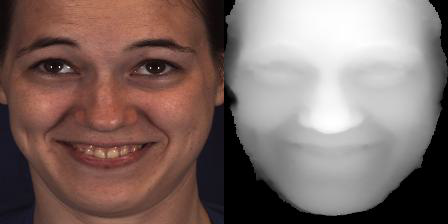}
     \end{subfigure}
     \hfill
     \begin{subfigure}[b]{0.28\textwidth}
         \centering
         \includegraphics[width=\textwidth]{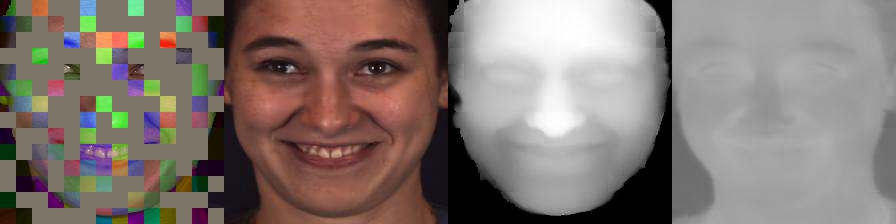}
     \end{subfigure}
     \hfill
     \begin{subfigure}[b]{0.28\textwidth}
         \centering
         \includegraphics[width=\textwidth]{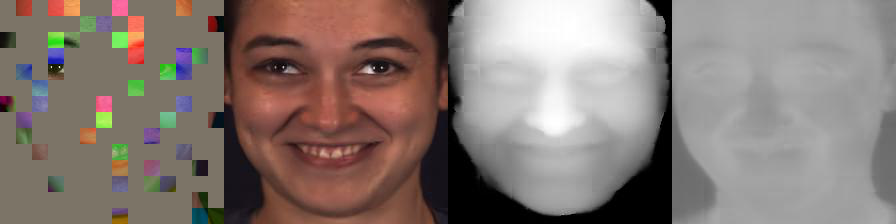}
     \end{subfigure}
     \hfill
     \begin{subfigure}[b]{0.28\textwidth}
         \centering
         \includegraphics[width=\textwidth]{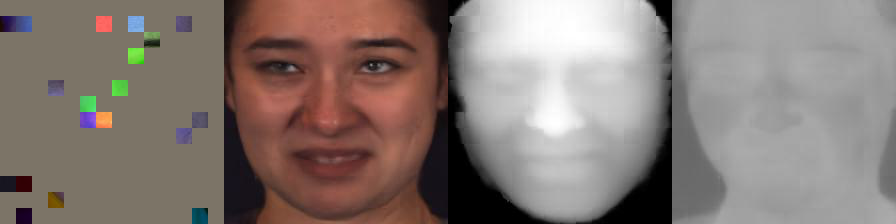}
     \end{subfigure}

     \begin{subfigure}[b]{0.14\textwidth}
         \centering
         \includegraphics[width=\textwidth]{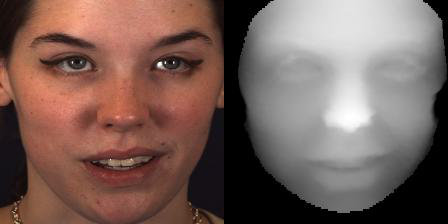}
     \end{subfigure}
     \hfill
     \begin{subfigure}[b]{0.28\textwidth}
         \centering
         \includegraphics[width=\textwidth]{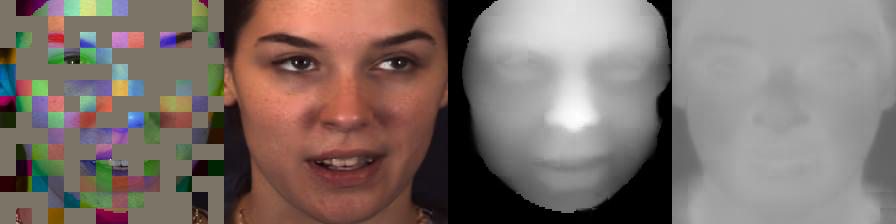}
     \end{subfigure}
     \hfill
     \begin{subfigure}[b]{0.28\textwidth}
         \centering
         \includegraphics[width=\textwidth]{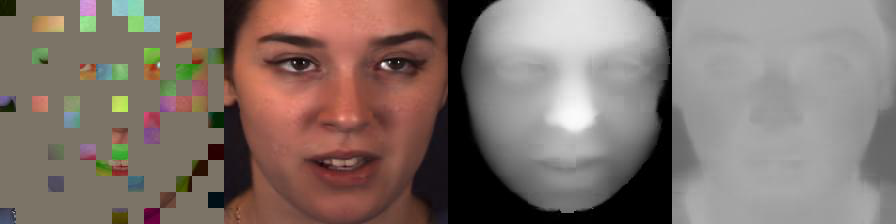}
     \end{subfigure}
     \hfill
     \begin{subfigure}[b]{0.28\textwidth}
         \centering
         \includegraphics[width=\textwidth]{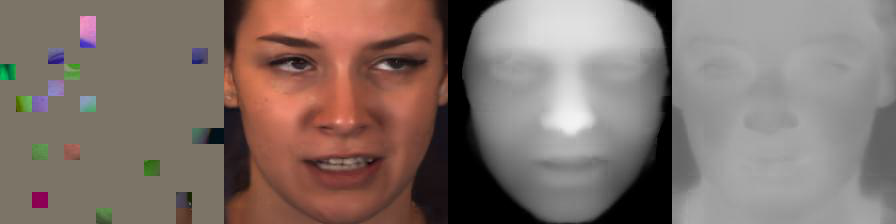}
     \end{subfigure}

     \begin{subfigure}[b]{0.14\textwidth}
         \centering
         \includegraphics[width=\textwidth]{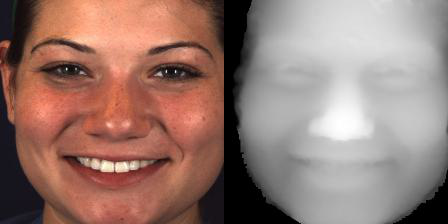}
     \end{subfigure}
     \hfill
     \begin{subfigure}[b]{0.28\textwidth}
         \centering
         \includegraphics[width=\textwidth]{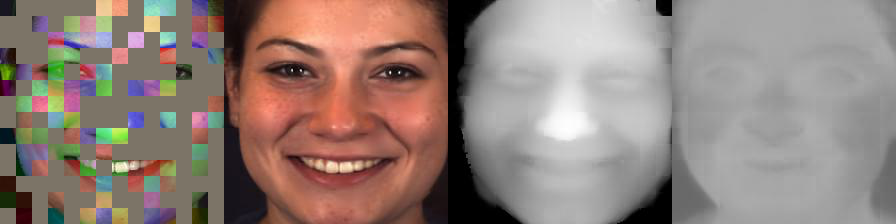}
     \end{subfigure}
     \hfill
     \begin{subfigure}[b]{0.28\textwidth}
         \centering
         \includegraphics[width=\textwidth]{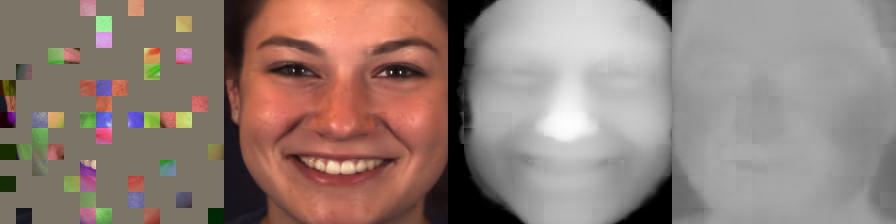}
     \end{subfigure}
     \hfill
     \begin{subfigure}[b]{0.28\textwidth}
         \centering
         \includegraphics[width=\textwidth]{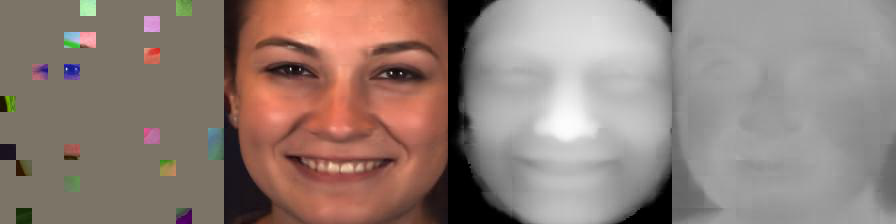}
     \end{subfigure}

     \begin{subfigure}[b]{0.14\textwidth}
         \centering
         \includegraphics[width=\textwidth]{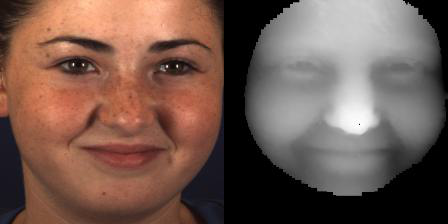}
     \end{subfigure}
     \hfill
     \begin{subfigure}[b]{0.28\textwidth}
         \centering
         \includegraphics[width=\textwidth]{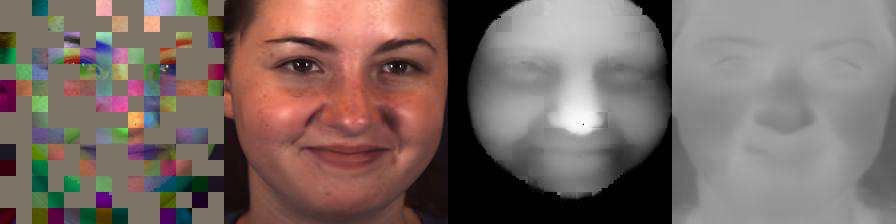}
     \end{subfigure}
     \hfill
     \begin{subfigure}[b]{0.28\textwidth}
         \centering
         \includegraphics[width=\textwidth]{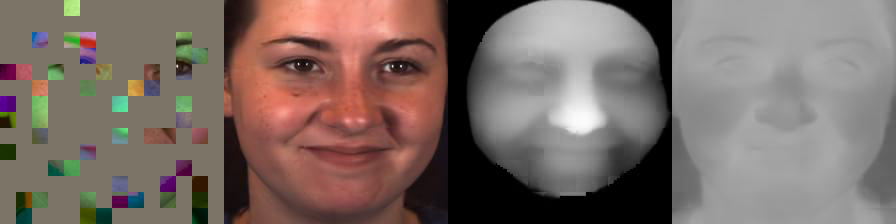}
     \end{subfigure}
     \hfill
     \begin{subfigure}[b]{0.28\textwidth}
         \centering
         \includegraphics[width=\textwidth]{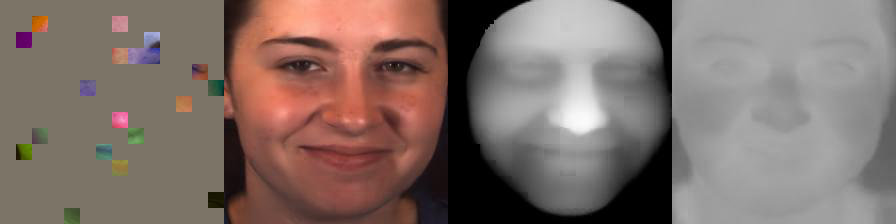}
     \end{subfigure}

     \begin{subfigure}[b]{0.14\textwidth}
         \centering
         \includegraphics[width=\textwidth]{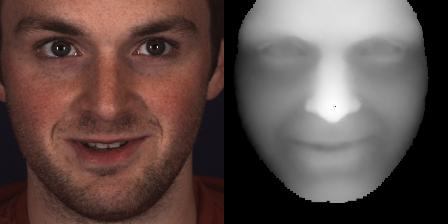}
     \end{subfigure}
     \hfill
     \begin{subfigure}[b]{0.28\textwidth}
         \centering
         \includegraphics[width=\textwidth]{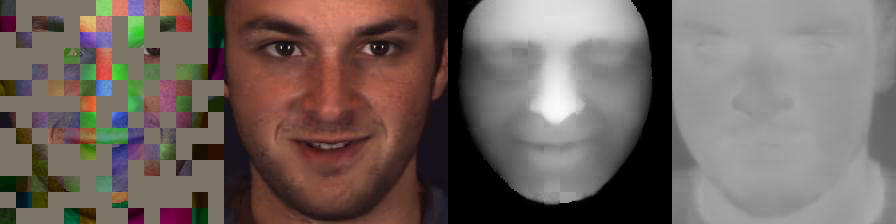}
     \end{subfigure}
     \hfill
     \begin{subfigure}[b]{0.28\textwidth}
         \centering
         \includegraphics[width=\textwidth]{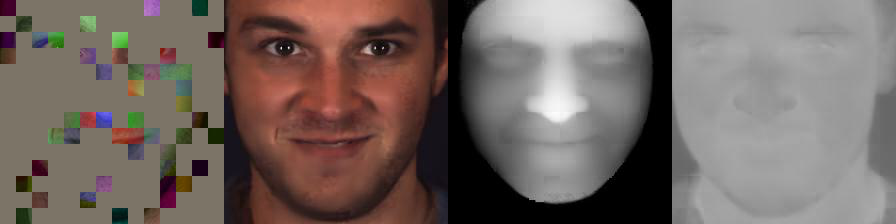}
     \end{subfigure}
     \hfill
     \begin{subfigure}[b]{0.28\textwidth}
         \centering
         \includegraphics[width=\textwidth]{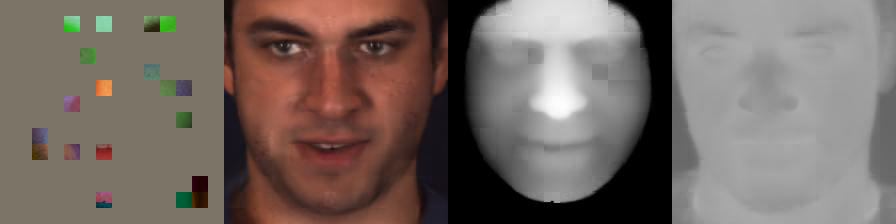}
     \end{subfigure}
     \begin{subfigure}[b]{0.14\textwidth}
         \centering
         \includegraphics[width=\textwidth]{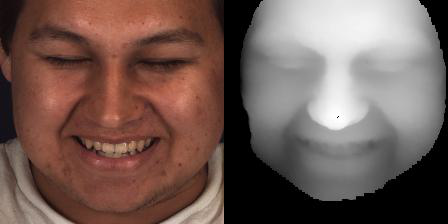}
     \end{subfigure}
     \hfill
     \begin{subfigure}[b]{0.28\textwidth}
         \centering
         \includegraphics[width=\textwidth]{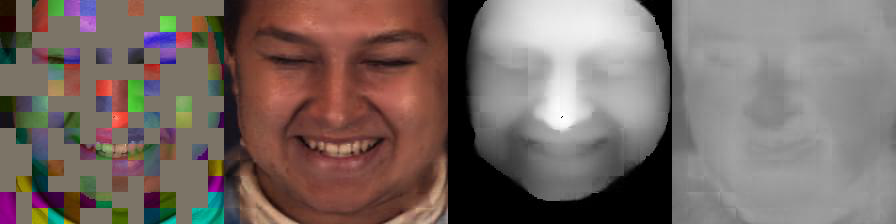}
     \end{subfigure}
     \hfill
     \begin{subfigure}[b]{0.28\textwidth}
         \centering
         \includegraphics[width=\textwidth]{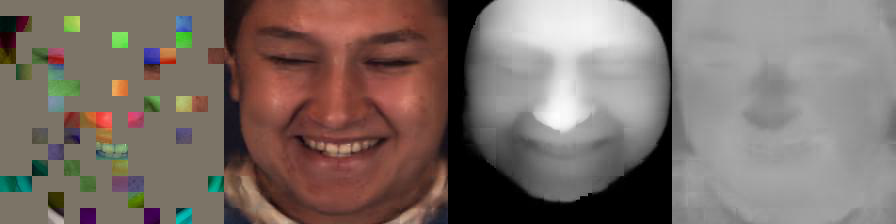}
     \end{subfigure}
     \hfill
     \begin{subfigure}[b]{0.28\textwidth}
         \centering
         \includegraphics[width=\textwidth]{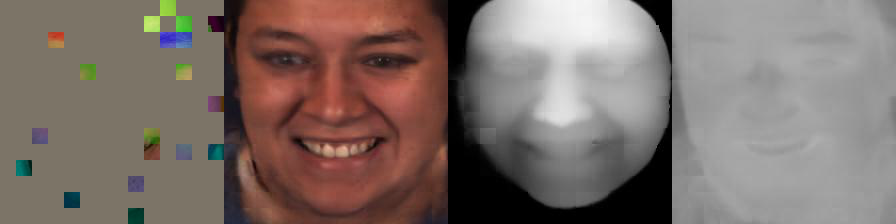}
     \end{subfigure}
     \begin{subfigure}[b]{0.14\textwidth}
         \centering
         \includegraphics[width=\textwidth]{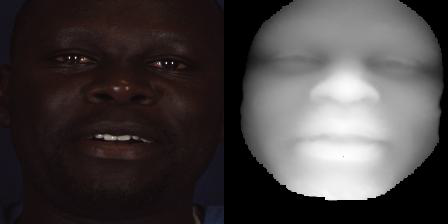}
     \end{subfigure}
     \hfill
     \begin{subfigure}[b]{0.28\textwidth}
         \centering
         \includegraphics[width=\textwidth]{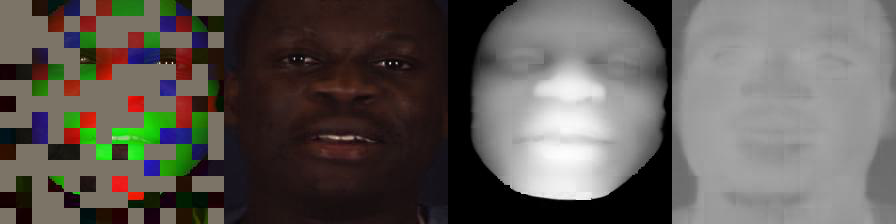}
     \end{subfigure}
     \hfill
     \begin{subfigure}[b]{0.28\textwidth}
         \centering
         \includegraphics[width=\textwidth]{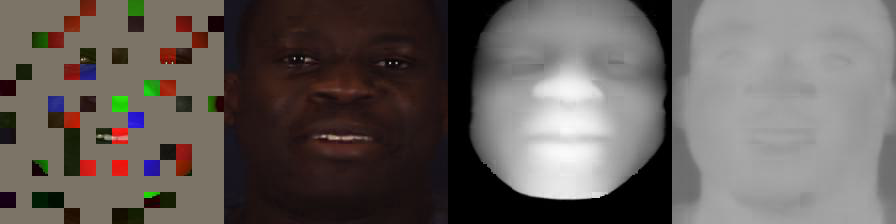}
     \end{subfigure}
     \hfill
     \begin{subfigure}[b]{0.28\textwidth}
         \centering
         \includegraphics[width=\textwidth]{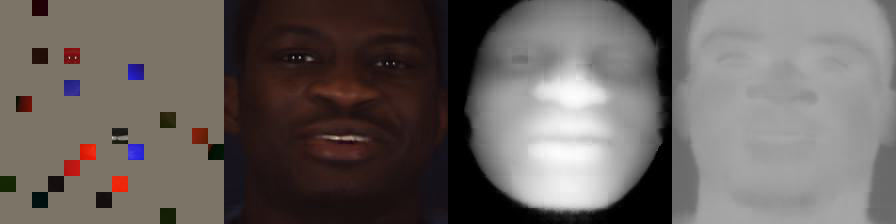}
     \end{subfigure}

    \begin{subfigure}[b]{0.14\textwidth}
         \centering
         \includegraphics[width=\textwidth]{supple_figures/BP4D/mask_0.5/sub1_origin.png}
     \end{subfigure}
     \hfill
     \begin{subfigure}[b]{0.28\textwidth}
         \centering
         \includegraphics[width=\textwidth]{supple_figures/BP4D/mask_0.5/sub1_mask.png}
     \end{subfigure}
     \hfill
     \begin{subfigure}[b]{0.28\textwidth}
         \centering
         \includegraphics[width=\textwidth]{supple_figures/BP4D/mask_0.75/sub1_mask.png}
     \end{subfigure}
     \hfill
     \begin{subfigure}[b]{0.28\textwidth}
         \centering
         \includegraphics[width=\textwidth]{supple_figures/BP4D/mask_0.9/sub1_mask.png}
     \end{subfigure}
     \begin{subfigure}[b]{0.14\textwidth}
         \centering
         \includegraphics[width=\textwidth]{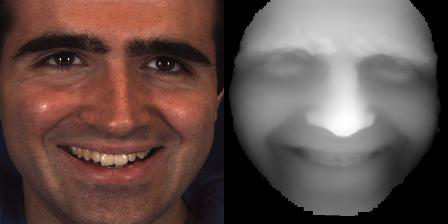}
     \end{subfigure}
     \hfill
     \begin{subfigure}[b]{0.28\textwidth}
         \centering
         \includegraphics[width=\textwidth]{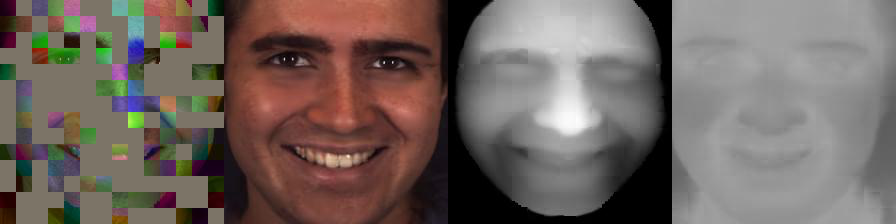}
     \end{subfigure}
     \hfill
     \begin{subfigure}[b]{0.28\textwidth}
         \centering
         \includegraphics[width=\textwidth]{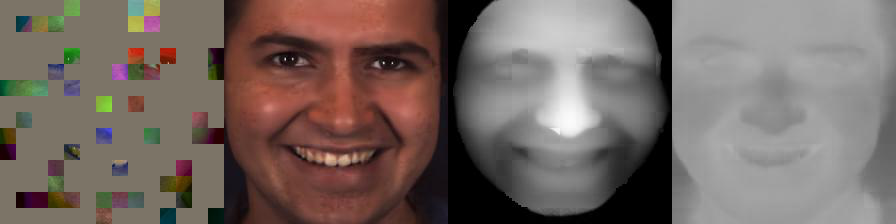}
     \end{subfigure}
     \hfill
     \begin{subfigure}[b]{0.28\textwidth}
         \centering
         \includegraphics[width=\textwidth]{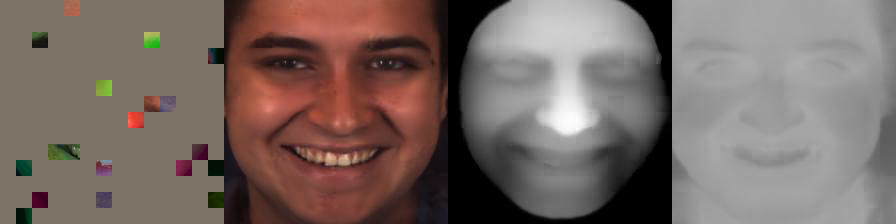}
     \end{subfigure}
     \begin{subfigure}[b]{0.14\textwidth}
         \centering
         \includegraphics[width=\textwidth]{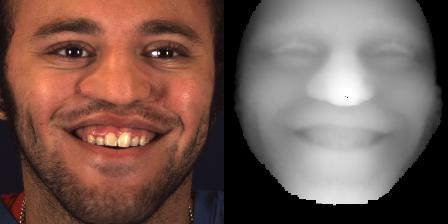}
     \end{subfigure}
     \hfill
     \begin{subfigure}[b]{0.28\textwidth}
         \centering
         \includegraphics[width=\textwidth]{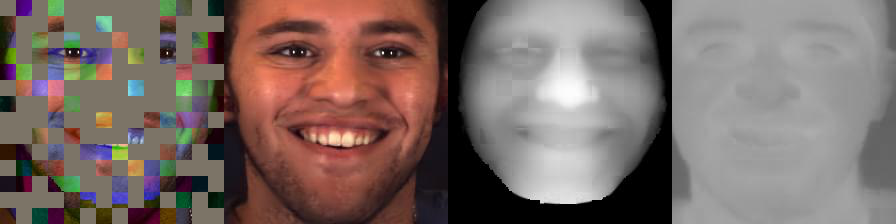}
     \end{subfigure}
     \hfill
     \begin{subfigure}[b]{0.28\textwidth}
         \centering
         \includegraphics[width=\textwidth]{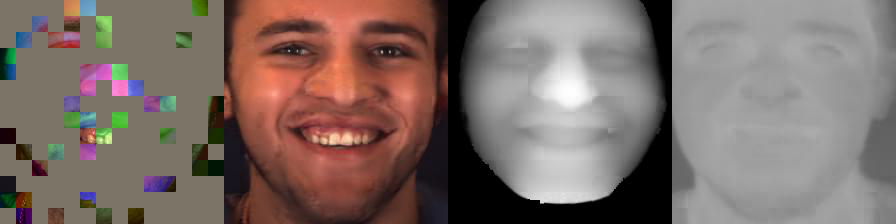}
     \end{subfigure}
     \hfill
     \begin{subfigure}[b]{0.28\textwidth}
         \centering
         \includegraphics[width=\textwidth]{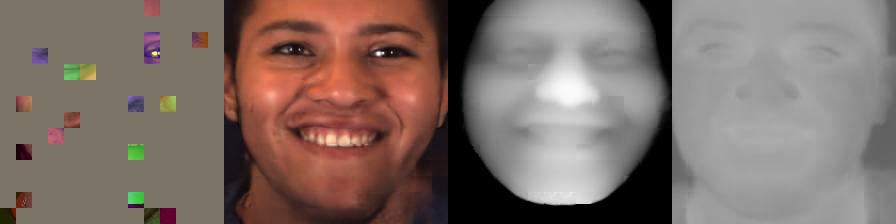}
     \end{subfigure}
     \begin{subfigure}[b]{0.14\textwidth}
         \centering
         \includegraphics[width=\textwidth]{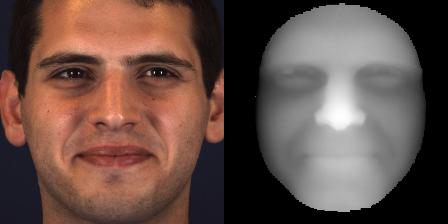}
     \end{subfigure}
     \hfill
     \begin{subfigure}[b]{0.28\textwidth}
         \centering
         \includegraphics[width=\textwidth]{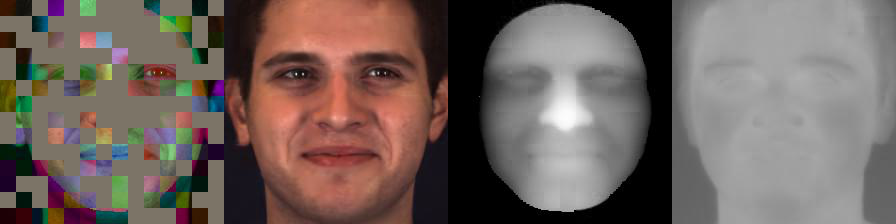}
     \end{subfigure}
     \hfill
     \begin{subfigure}[b]{0.28\textwidth}
         \centering
         \includegraphics[width=\textwidth]{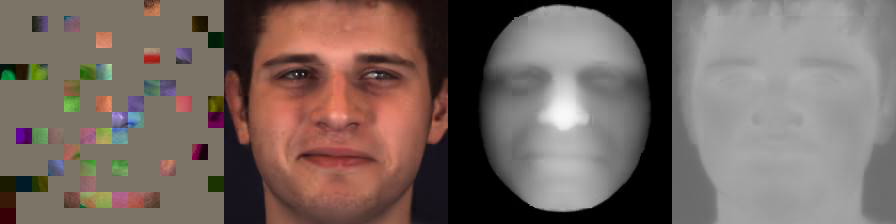}
     \end{subfigure}
     \hfill
     \begin{subfigure}[b]{0.28\textwidth}
         \centering
         \includegraphics[width=\textwidth]{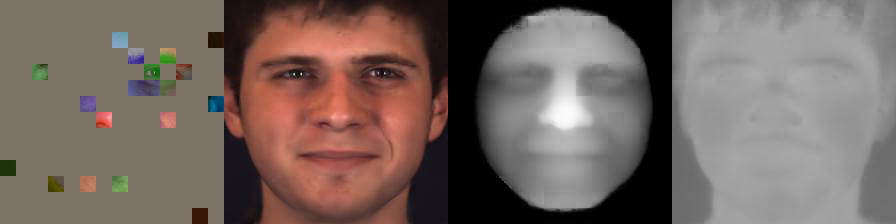}
     \end{subfigure}
     \begin{subfigure}[b]{0.14\textwidth}
         \centering
         \includegraphics[width=\textwidth]{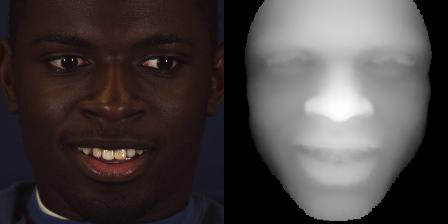}
     \end{subfigure}
     \hfill
     \begin{subfigure}[b]{0.28\textwidth}
         \centering
         \includegraphics[width=\textwidth]{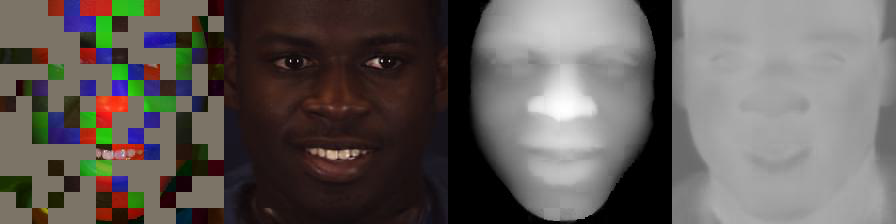}
     \end{subfigure}
     \hfill
     \begin{subfigure}[b]{0.28\textwidth}
         \centering
         \includegraphics[width=\textwidth]{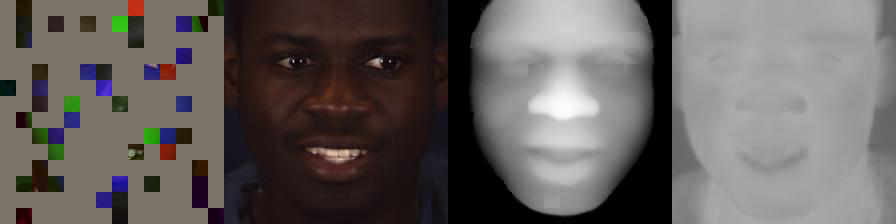}
     \end{subfigure}
     \hfill
     \begin{subfigure}[b]{0.28\textwidth}
         \centering
         \includegraphics[width=\textwidth]{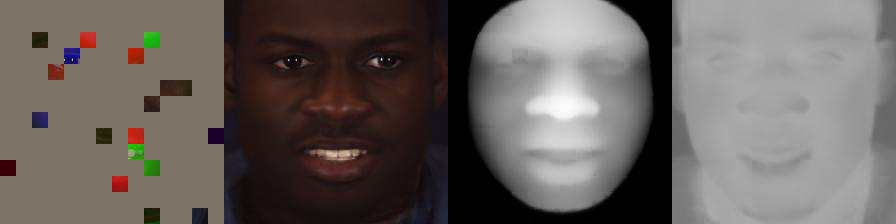}
     \end{subfigure}
     \begin{subfigure}[b]{0.14\textwidth}
         \centering
         \includegraphics[width=\textwidth]{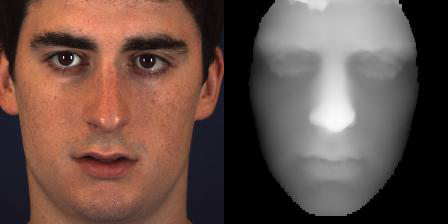}
         \caption{origin}
     \end{subfigure}
     \hfill
     \begin{subfigure}[b]{0.28\textwidth}
         \centering
         \includegraphics[width=\textwidth]{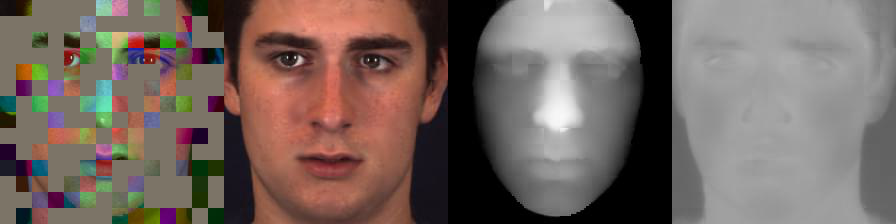}
         \caption{mask: 50\%}
     \end{subfigure}
     \hfill
     \begin{subfigure}[b]{0.28\textwidth}
         \centering
         \includegraphics[width=\textwidth]{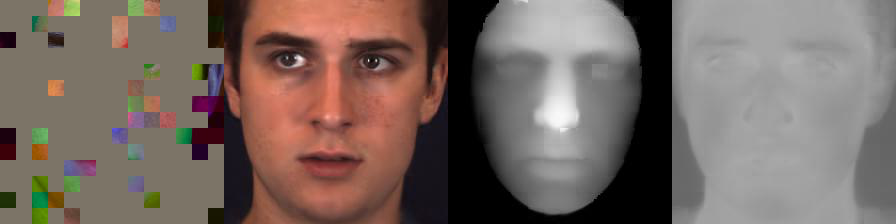}
         \caption{mask: 75\%}
     \end{subfigure}
     \hfill
     \begin{subfigure}[b]{0.28\textwidth}
         \centering
         \includegraphics[width=\textwidth]{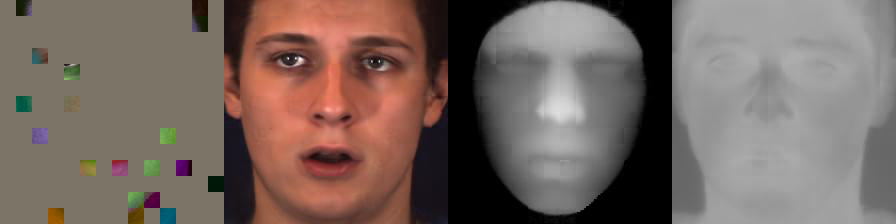}
         \caption{mask: 90\%}
     \end{subfigure}
\caption{
Visualization of the reconstructed images on BP4D with different mask ratios is performed using a reconstruction model trained on BP4D+. 
Images in (b, c, and d) represent the masked channel-mixing image, reconstructed RGB, reconstructed Depth, and reconstructed Thermal, respectively.
}
\label{fig:mask_samples_bp4d}
\end{figure*}

\begin{figure*}[th]
     \centering
     \begin{subfigure}[b]{0.33\textwidth}
         \centering
         \includegraphics[width=\textwidth]{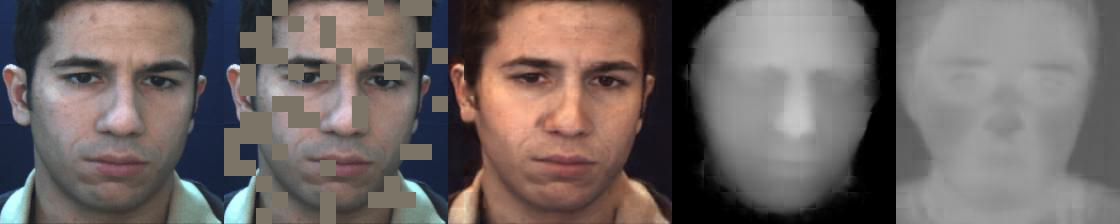}
     \end{subfigure}
     \hfill
     \begin{subfigure}[b]{0.33\textwidth}
         \centering
         \includegraphics[width=\textwidth]{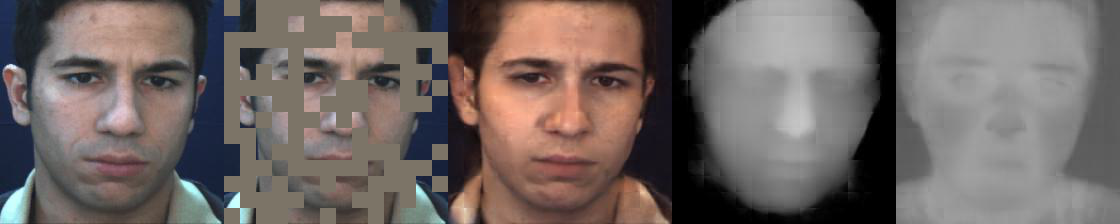}
     \end{subfigure}
     \hfill
     \begin{subfigure}[b]{0.33\textwidth}
         \centering
         \includegraphics[width=\textwidth]{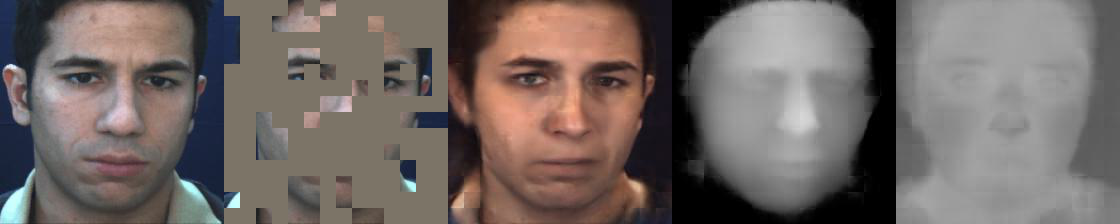}
     \end{subfigure}
     \begin{subfigure}[b]{0.33\textwidth}
         \centering
         \includegraphics[width=\textwidth]{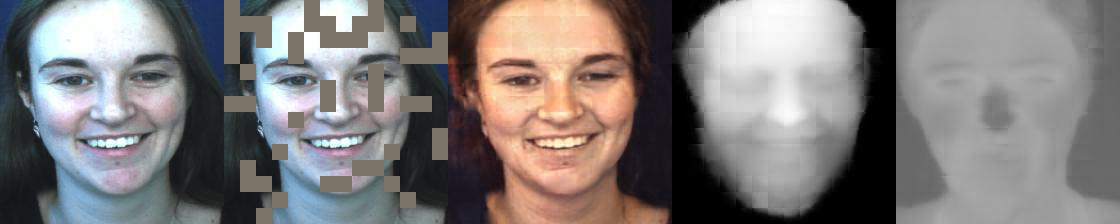}
     \end{subfigure}
     \hfill
     \begin{subfigure}[b]{0.33\textwidth}
         \centering
         \includegraphics[width=\textwidth]{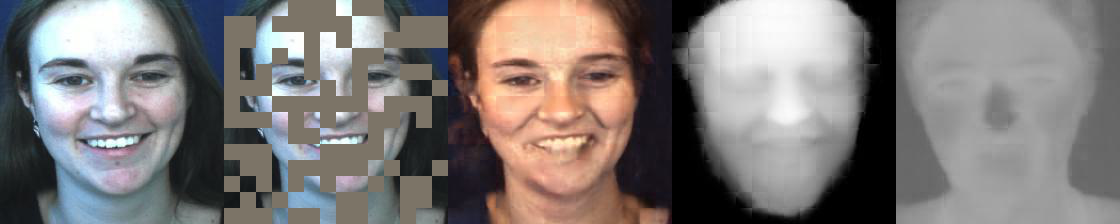}
     \end{subfigure}
     \hfill
     \begin{subfigure}[b]{0.33\textwidth}
         \centering
         \includegraphics[width=\textwidth]{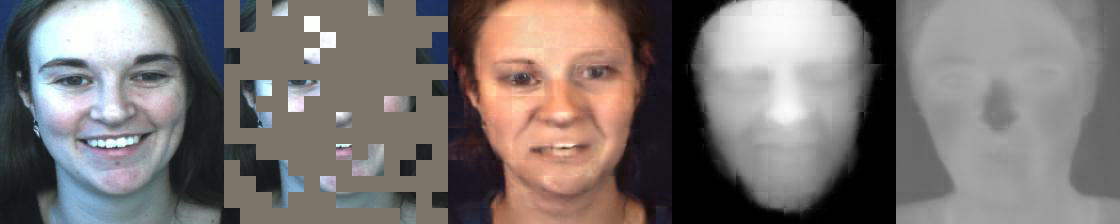}
     \end{subfigure}
     \begin{subfigure}[b]{0.33\textwidth}
         \centering
         \includegraphics[width=\textwidth]{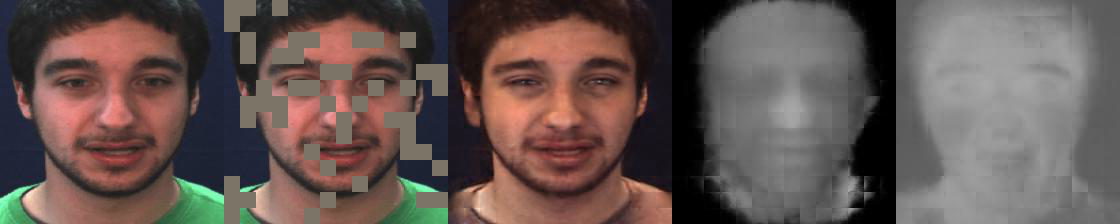}
     \end{subfigure}
     \hfill
     \begin{subfigure}[b]{0.33\textwidth}
         \centering
         \includegraphics[width=\textwidth]{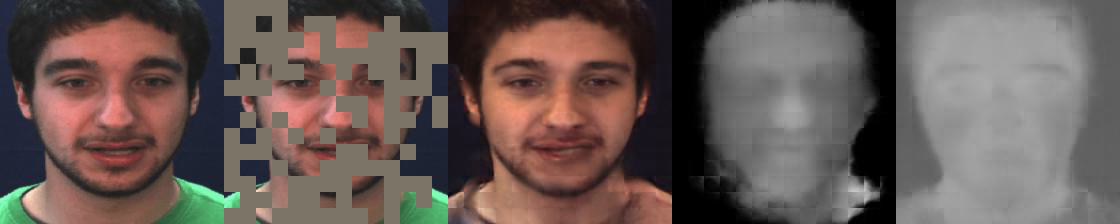}
     \end{subfigure}
     \hfill
     \begin{subfigure}[b]{0.33\textwidth}
         \centering
         \includegraphics[width=\textwidth]{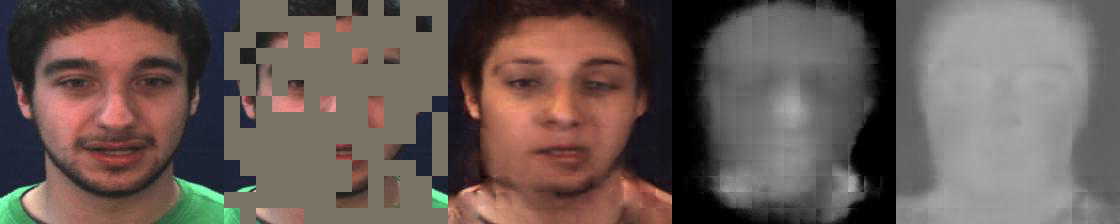}
     \end{subfigure}
     \begin{subfigure}[b]{0.33\textwidth}
         \centering
         \includegraphics[width=\textwidth]{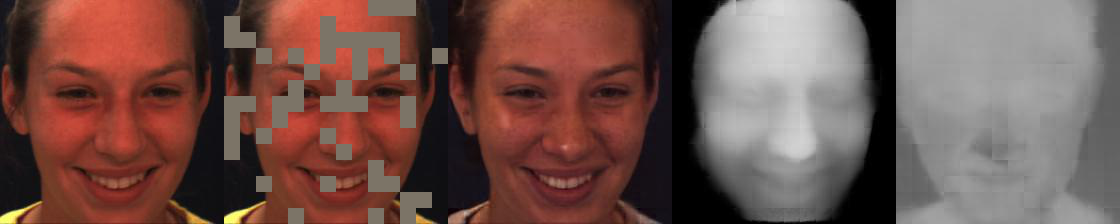}
     \end{subfigure}
     \hfill
     \begin{subfigure}[b]{0.33\textwidth}
         \centering
         \includegraphics[width=\textwidth]{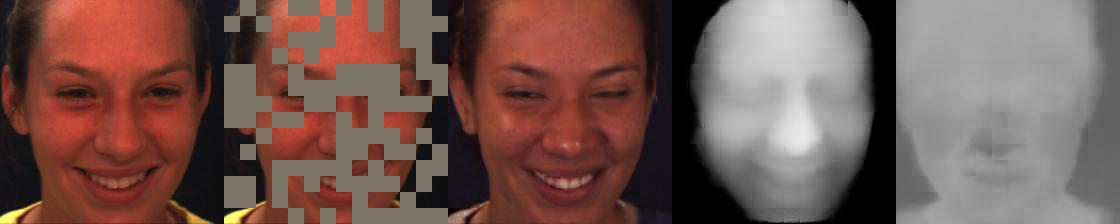}
     \end{subfigure}
     \hfill
     \begin{subfigure}[b]{0.33\textwidth}
         \centering
         \includegraphics[width=\textwidth]{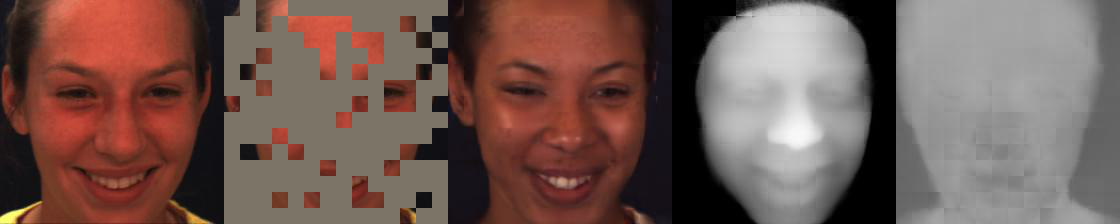}
     \end{subfigure}
     \begin{subfigure}[b]{0.33\textwidth}
         \centering
         \includegraphics[width=\textwidth]{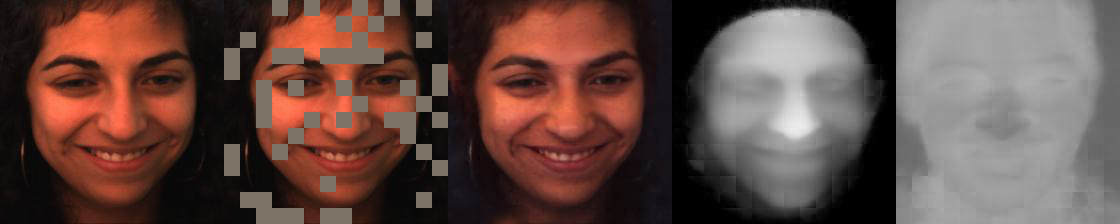}
     \end{subfigure}
     \hfill
     \begin{subfigure}[b]{0.33\textwidth}
         \centering
         \includegraphics[width=\textwidth]{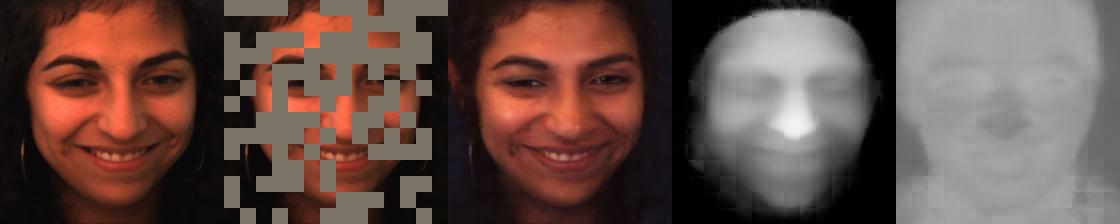}
     \end{subfigure}
     \hfill
     \begin{subfigure}[b]{0.33\textwidth}
         \centering
         \includegraphics[width=\textwidth]{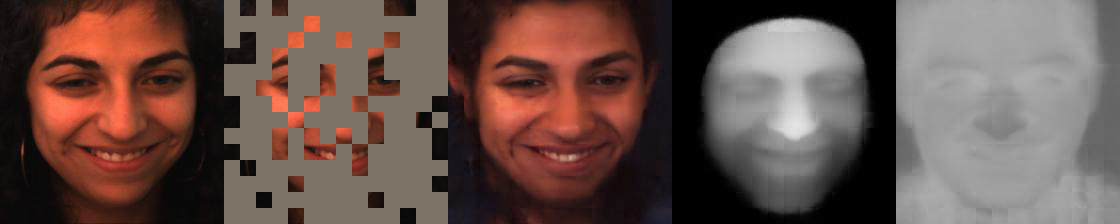}
     \end{subfigure}
     \begin{subfigure}[b]{0.33\textwidth}
         \centering
         \includegraphics[width=\textwidth]{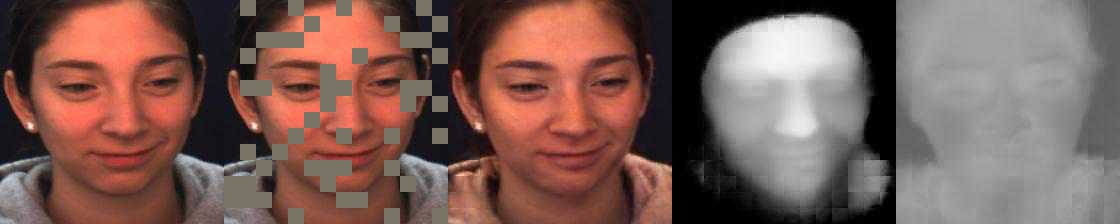}
     \end{subfigure}
     \hfill
     \begin{subfigure}[b]{0.33\textwidth}
         \centering
         \includegraphics[width=\textwidth]{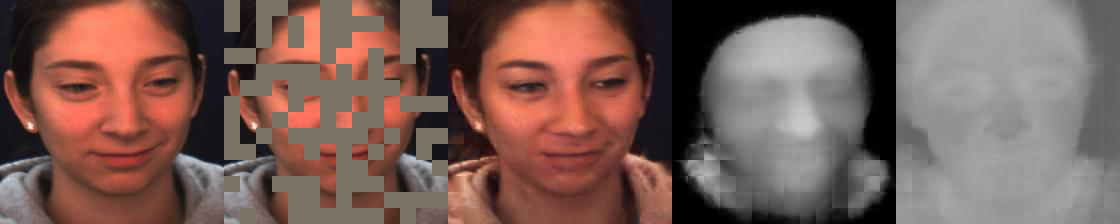}
     \end{subfigure}
     \hfill
     \begin{subfigure}[b]{0.33\textwidth}
         \centering
         \includegraphics[width=\textwidth]{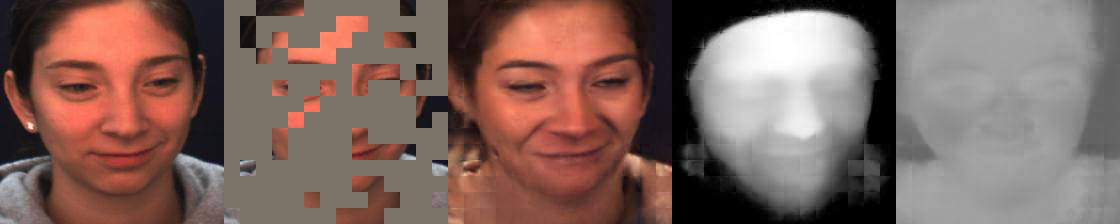}
     \end{subfigure}
     \begin{subfigure}[b]{0.33\textwidth}
         \centering
         \includegraphics[width=\textwidth]{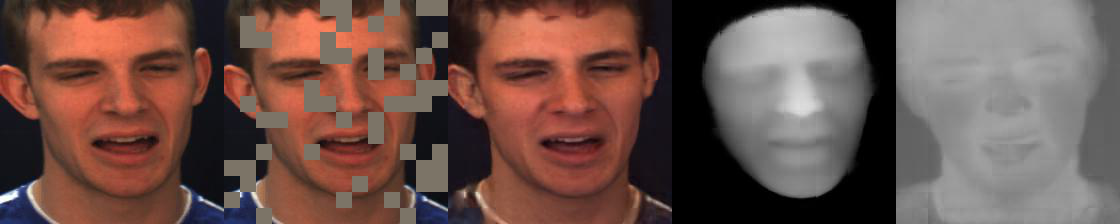}
     \end{subfigure}
     \hfill
     \begin{subfigure}[b]{0.33\textwidth}
         \centering
         \includegraphics[width=\textwidth]{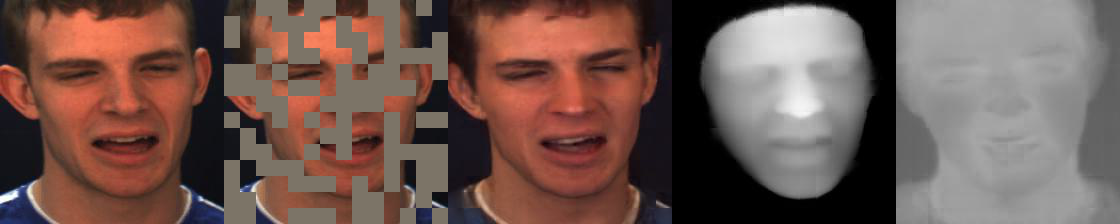}
     \end{subfigure}
     \hfill
     \begin{subfigure}[b]{0.33\textwidth}
         \centering
         \includegraphics[width=\textwidth]{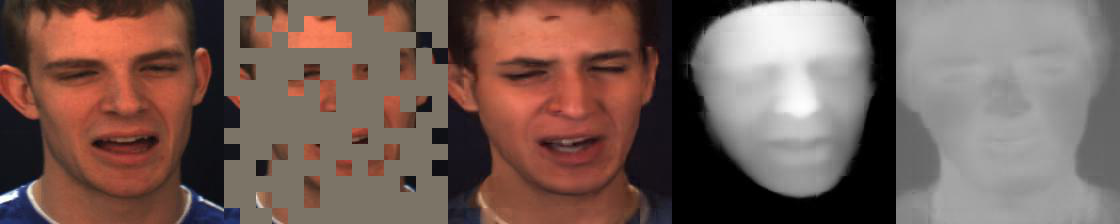}
     \end{subfigure}
     \begin{subfigure}[b]{0.33\textwidth}
         \centering
         \includegraphics[width=\textwidth]{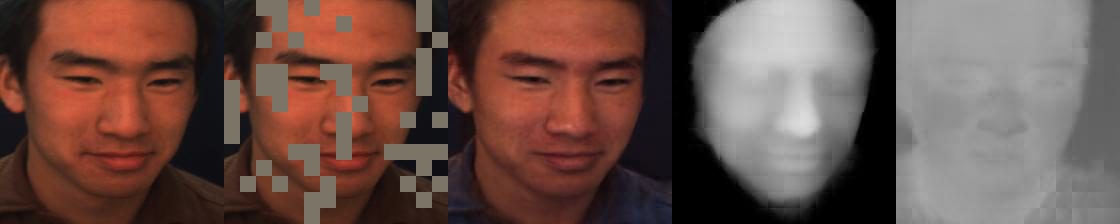}
     \end{subfigure}
     \hfill
     \begin{subfigure}[b]{0.33\textwidth}
         \centering
         \includegraphics[width=\textwidth]{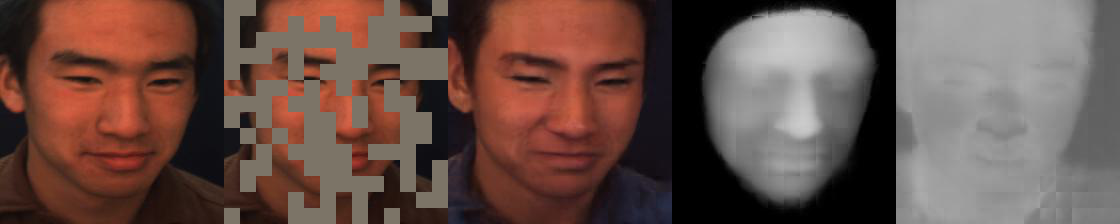}
     \end{subfigure}
     \hfill
     \begin{subfigure}[b]{0.33\textwidth}
         \centering
         \includegraphics[width=\textwidth]{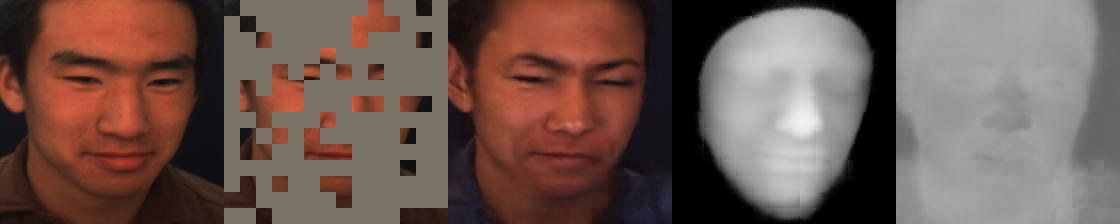}
     \end{subfigure}
     \begin{subfigure}[b]{0.33\textwidth}
         \centering
         \includegraphics[width=\textwidth]{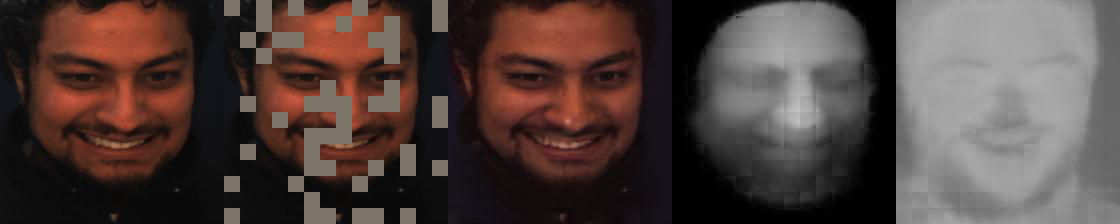}
         \caption{mask: 25\%}
     \end{subfigure}
     \hfill
     \begin{subfigure}[b]{0.33\textwidth}
         \centering
         \includegraphics[width=\textwidth]{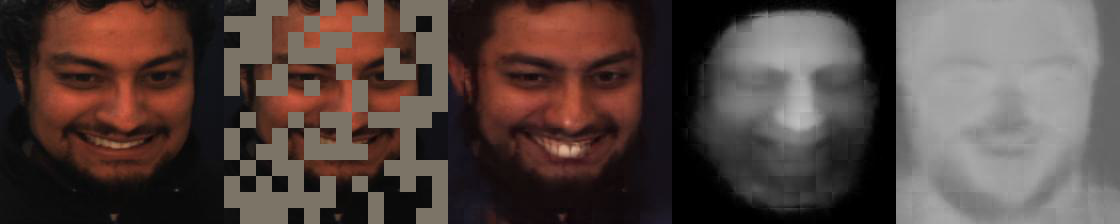}
         \caption{mask: 50\%}
     \end{subfigure}
     \hfill
     \begin{subfigure}[b]{0.33\textwidth}
         \centering
         \includegraphics[width=\textwidth]{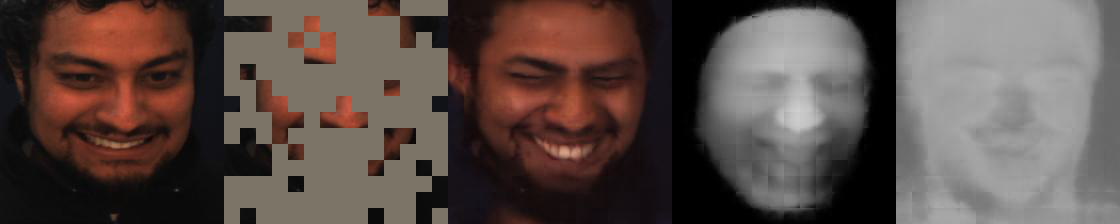}
         \caption{mask: 75\%}
     \end{subfigure}
\caption{
Visualization of the reconstructed images on DISFA with different mask ratios is performed using a reconstruction model trained on BP4D+. 
Images in each cell represent the original RGB, masked image, reconstructed RGB, reconstructed Depth, and reconstructed Thermal, respectively.
}
\label{fig:mask_samples_disfa}
\end{figure*}

More visualization examples on BP4D and DISFA are shown in Figure~\ref{fig:mask_samples_bp4d} and Figure~\ref{fig:mask_samples_disfa}.


